\documentclass[runningheads]{llncs}
\usepackage{graphicx}

\usepackage{tikz}
\usepackage{comment}
\usepackage{amsmath,amssymb} 
\usepackage{color}

\usepackage[pagebackref=true,breaklinks=true,letterpaper=true,colorlinks,bookmarks=false]{hyperref}

\usepackage{booktabs}
\usepackage{enumitem}

\usepackage{wrapfig}
\usepackage{bm}
\usepackage{algorithm}
\usepackage{algorithmic}
\usepackage{listings}
\usepackage{caption}
\usepackage{float}
\usepackage{array}
\usepackage{multirow}

\usepackage{colortbl}
\definecolor{mygraylite}{gray}{.91}
\definecolor{mygray}{gray}{.89}
\definecolor{darkergreen}{RGB}{21, 152, 56}

\definecolor{darkergreen}{RGB}{21, 152, 56}
\definecolor{red2}{RGB}{252, 54, 65}

\begin{document}
	\pagestyle{headings}
	\mainmatter

	\title{Sliced Recursive Transformer} 
	
	
		\titlerunning{Sliced Recursive Transformer}
		%
		\author{Zhiqiang Shen\inst{1,2,3} \and
			Zechun Liu\inst{2,4} \and
			Eric Xing\inst{1,3}}
		%
		\authorrunning{Zhiqiang Shen et al.}
		%
		\institute{Carnegie Mellon University, Pittsburgh, USA \and
			Hong Kong University of Science and Technology, Hong Kong, China \and
			Mohamed bin Zayed University of Artificial Intelligence, Abu Dhabi, UAE \and
			Reality Labs, Meta Inc.
			\email{zhiqiangshen@cse.ust.hk,zechunliu@fb.com,epxing@cs.cmu.edu}
		}
\newcommand{\methodname}{dense convolutional  network}
\newcommand{\methodnamecap}{Dense Convolutional Network}
\newcommand{\methodnameshort}{DenseNet}
\newcommand{\methodnameshorts}{DenseNets}
\newcommand{\methodblock}{dense block}
\newcommand{\methodblockcap}{Dense Block}

\newcommand{\regmethodname}{feature drop}
\newcommand{\regmethodnamecap}{Feature Drop}

\newcommand{\stepsizename}{growth rate}

\newcommand{\conv}[1]{$\left[\begin{array}{ll} \text{1}\times \text{1} \text{ conv}\\ \text{3}\times \text{3} \text{ conv} \end{array}\right] \times \text{#1}$}

\newcommand{\ret}[5]{$\left[\left[\begin{array}{ll} \text{ \ \ \ #3-dim} \text{ MHSA}\\ \text{#4$\times$}\text{FFN} / \text{#5$\times$}\text{NLL}  \end{array}\right] \times \text{#1}\right] \times \text{#2}$}

\newcommand{\cross}[1]{#1 $\times$ #1}

\newcommand{\feati}{x_i}
\newcommand{\clsfeati}{y_i}
\newcommand{\featk}{x_k}
\newcommand{\clsfeatk}{y_k}
\newcommand{\loss}{L}
\newcommand{\featL}{x_L}
\newcommand{\clsfeat}{y}
\newcommand{\anyxs}{\ensuremath{\mathbf{x}}}
\newcommand{\anyys}{\ensuremath{\mathbf{y}}}

\newcommand{\bx}{\ensuremath{\mathbf{x}}}

\newcommand{\sourcexs}{\ensuremath{\mathbf{x^\mathcal{S}}}}
\newcommand{\sourceys}{\ensuremath{\mathbf{y^\mathcal{S}}}}

\newcommand{\targetxs}{\ensuremath{\mathbf{x^\mathcal{T}}}}
\newcommand{\targetys}{\ensuremath{\mathbf{y^\mathcal{T}}}}
\newcommand{\pseudotargetys}{\ensuremath{\mathbf{\hat{y}^\mathcal{T}}}}

	\maketitle
	
	\begin{abstract}
		We present a neat yet effective recursive operation on vision transformers that can improve parameter utilization without involving additional parameters. This is achieved by sharing weights across depth of transformer networks. 
		The proposed method can obtain a substantial gain ($\sim$2\%) simply using na\"ive recursive operation, requires no special or sophisticated knowledge for designing principles of networks, and introduces minimal computational overhead to the training procedure. To reduce the additional computation caused by recursive operation while maintaining the superior accuracy, we propose an approximating method through multiple sliced group self-attentions across recursive layers which can reduce the cost consumption by 10$\sim$30\% without sacrificing performance. We call our model {\bf S}liced {\bf Re}cursive {\bf T}ransformer (SReT), a novel and parameter-efficient vision transformer design that is compatible with a broad range of other designs for efficient ViT architectures. Our best model establishes significant improvement on ImageNet-1K over state-of-the-art methods while containing fewer parameters. The proposed weight sharing mechanism by sliced recursion structure allows us to build a transformer with more than 100 or even 1000 shared layers with ease while keeping a compact size (13$\sim$15M), to avoid optimization difficulties when the model is too large. The flexible scalability has shown great potential for scaling up models and constructing extremely deep vision transformers. Code is available at \url{https://github.com/szq0214/SReT}.
	\end{abstract}
	
	\section{Introduction}\label{introduction}
	
	\begin{wrapfigure}{r}{7.3cm}\vspace{-0.32in}
		\includegraphics[width=0.48\linewidth]{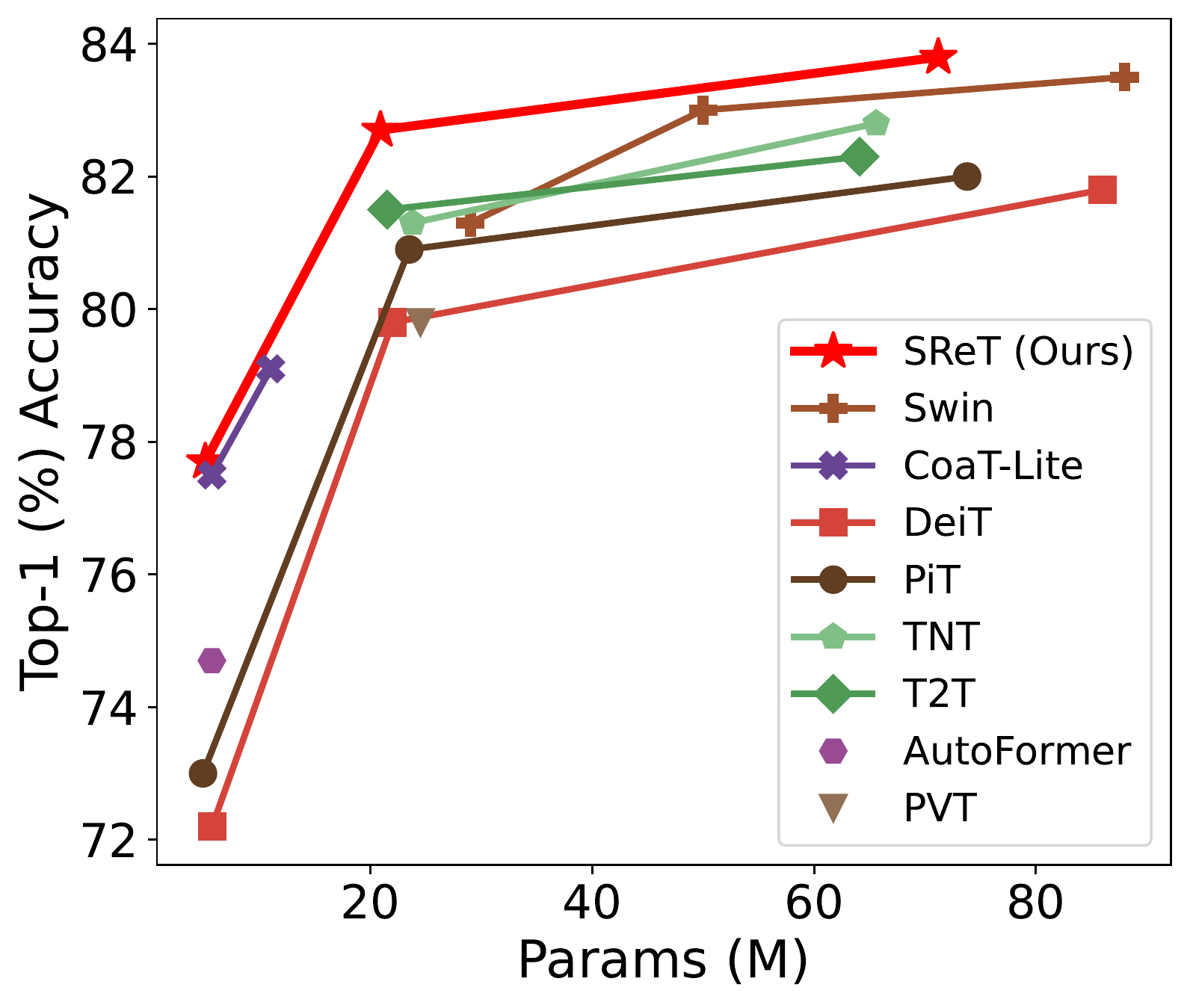}
		\includegraphics[width=0.48\linewidth]{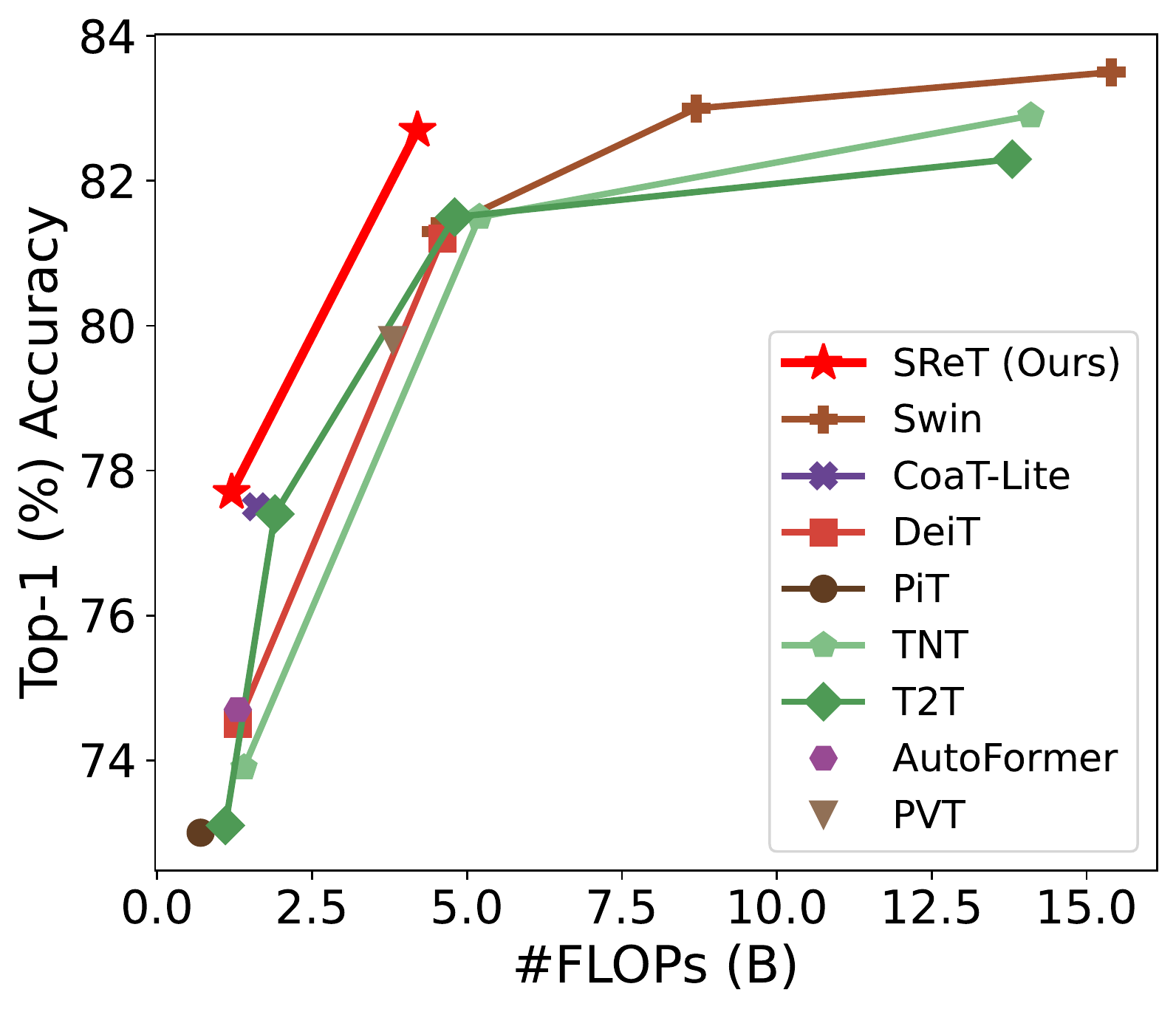}
		\vspace{-0.1in}\caption{Params/FLOPs vs. ImageNet-1K Acc.}
		\label{fig:params_flops}\vspace{-0.3in}
	\end{wrapfigure}
	
\begin{figure*}[t]
		\begin{minipage}{0.45\textwidth}\vspace{0.1in}\centering
			\includegraphics[width=0.85\linewidth]{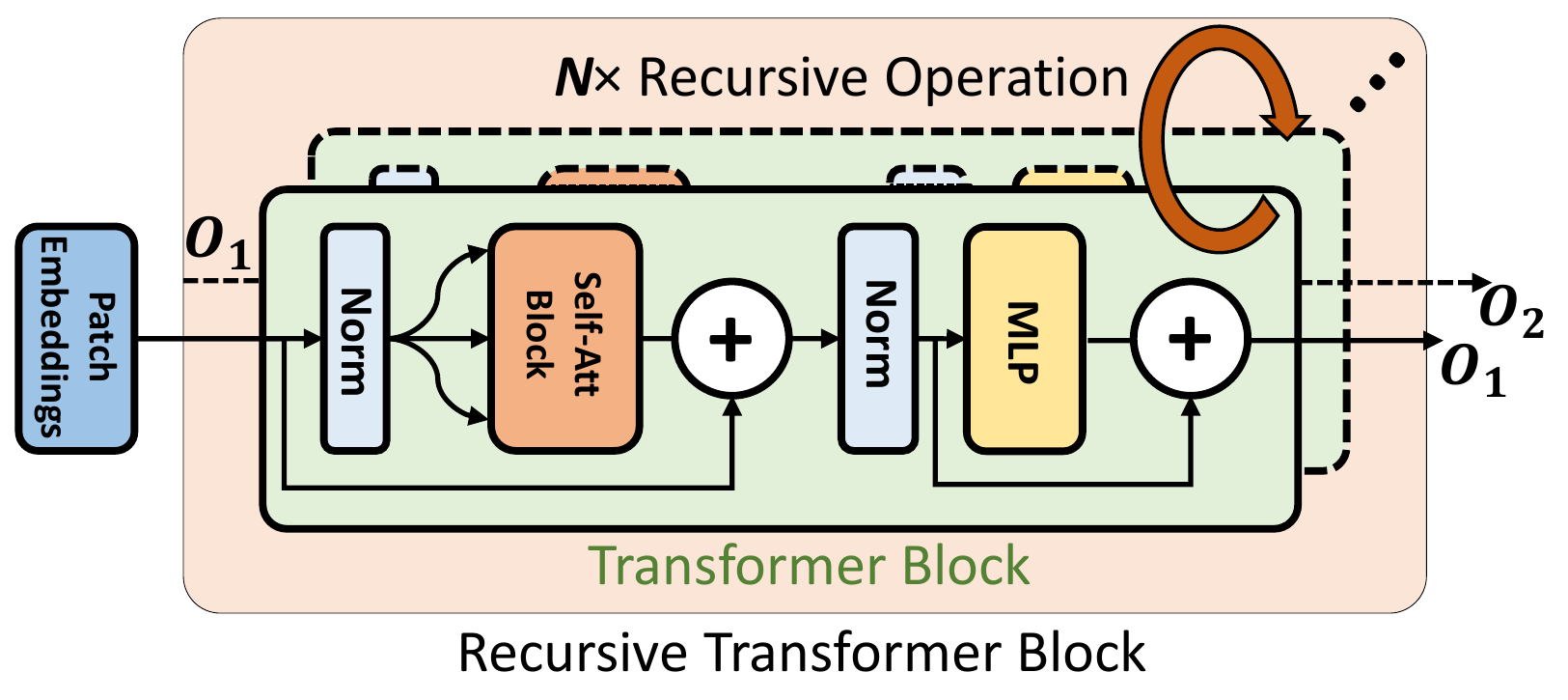}
			\caption{Atomic Recursive Operation.}
			\label{fig:expr:combinatorial_rec}
		\end{minipage}\hspace{0.1in}
		\begin{minipage}{0.5\textwidth}
			\setlength{\tabcolsep}{2pt}
			\captionof{table}{Results using different numbers $N$ of na\"ive recursive operation on ImageNet-1K dataset.}
			\label{tab:increase-table}\vspace{-0.06in}
			\resizebox{.99\textwidth}{!}{
				\centering
				\begin{tabular}{l|c|c|c}
					\toprule[1.1pt]
					Method        & Layers & \#Params (M) & Top-1 Acc. (\%) \\ \hline
					DeiT-Tiny~\cite{touvron2020training}  &  12  &    5.7    &  72.2     \\ \hline \hline
					+ 1$\times$ na\"ive recursion & 24  &    5.7    &   74.0      \\ \hline
					+ 2$\times$ na\"ive recursion  & 36  &   5.7   &   74.1        \\ \hline
					+ 3$\times$ na\"ive recursion  & 48 &   5.7   &   73.6       \\ 
					\bottomrule[1.1pt] 
			\end{tabular}}
		\end{minipage}\vspace{-0.06in}
	\end{figure*}

	The architectures of transformer have achieved substantively breakthroughs recently in the fields of natural language processing (NLP)~\cite{vaswani2017attention}, computer vision (CV)~\cite{dosovitskiy2021an} and speech~\cite{8462506,wang2021transformer}. In the vision area, Dosovitskiy et al.~\cite{dosovitskiy2021an} introduced a vision transformer (ViT) model that splits a raw image into a patch sequence as input, and they directly adopt transformer model~\cite{vaswani2017attention} for the image classification task. ViT achieved impressive results and has inspired many follow-up works. However, the benefits of a transformer often come with a large number of parameters and computational cost and it is always of great challenge to achieve the optimal trade-off between the accuracy and model complexity. In this work, we are motivated by the following question: { How can we improve the parameter utilization of a vision transformer, i.e., the representation ability without increasing the model size?} We observe recursive operation, as shown in Fig.~\ref{fig:expr:combinatorial_rec}, is a simple yet effective way to achieve this purpose. Our recursion-based vision transformer models significantly outperform state-of-the-art approaches while containing fewer parameters and FLOPs, as illustrated in Fig.~\ref{fig:params_flops}.
	
	Intrinsically, the classifier requires high-level abstracted features from the neural network to perform accurate classification, while the extraction of these features often requires multiple layers and deeper networks. This introduces parameter overhead into the model. Our motivation of this work stems from an interesting phenomenon of latent representation visualization. We observed that in the deep vision transformer network, the weights and activations of adjacent layers are similar with not much difference (a similar phenomenon is also discovered in~\cite{zhou2021deepvit}), which means they can be reused. The transformer with a fixed stack of distinct layers loses the inductive bias in the recurrent neural network which inspires us to share those weights in a recursive manner, forming an iterative or recursive vision transformer. Recursion can help extract stronger features without the need of increasing the parameters, and further improve the accuracy. In addition, this weight reuse or sharing strategy partially regularizes the training process by reducing the number of parameters to avoid overfitting and ill-convergence challenges, which will be discussed in the later sections.
	
	\noindent{\textbf{Why do we need to introduce {\em sliced recursion}, i.e., the group self-attention, into transformers? (advantages and drawbacks)}} We usually push towards perfection on weight utilization of a network under a bounded range of parameters, thus, it can be used practically in the resource-limited circumstances like embedded devices. Recursion is a straightforward way to {\em compress} the feature representation in a cyclic scheme. The recursive neural networks also allow the branching of connections and structures with hierarchies. We found that it is intriguingly crucial for learning better representations on vision data in a hierarchical manner, as we will introduce in Fig.~\ref{fig:feature_maps_vis} of our experiments. Also, even the most simplistic recursive operation still improves the compactness of utilizing parameters without requiring to modify the transformer block structure, unlike others~\cite{srinivas2021bottleneck,yuan2021tokens,heo2021rethinking,wang2021pyramid,wu2021cvt,liu2021swin,li2021localvit,xu2021co}, that add more parameters or involve additional fine-grained information from input~\cite{han2021transformer}. However, such a recursion will incur more computational cost by its loops, namely, it {\em sacrifices the executing efficiency for better parameter representation utilization}. To address this shortcoming, we propose an approximating method for global self-attention through decomposing into multiple sliced group self-attentions across recursive layers, meanwhile, enjoying similar FLOPs and better representations, we also apply the spatial pyramid design to reduce the complexity of the network.
	
	\noindent{\textbf{Feed-forward Networks, Recurrent Neural Networks and Recursive Neural Networks.}} 
Feed-forward networks, such as CNNs and transformers, are directed acyclic graphs (DAG), so the information path in the feed-forward processing is unidirectional. Recurrent networks (RNNs) are usually developed to process the time-series and other sequential data, and predict using current input and past memory. Recursive network is a less frequently used term compared to other two counterparts. Recursive refers to repeating or reusing a certain piece of a network\footnote{In a broader sense, the recurrent neural network is a type of recursive neural network.}. Different from RNNs that repeat the same block throughout the whole network, recursive network selectively repeats critical blocks for particular purposes. The recursive transformer iteratively refines its representations for all patches in the sequence. We found that, through the designed recursion into the feed-forward transformer, we can dramatically enhance feature representation especially for structured data without including additional parameters.

	The strong experimental results show that integrating the proposed sliced recursive operation in the transformer strike a competitive trade-off among accuracy, model size and complexity. To the best of our knowledge, there are barely existing works studying the effectiveness of recursive operation in vision transformers and proposing the approximation of self-attention method for reducing the complexity of recursive operation. We have done extensive experiments to derive a set of guidelines for the new design on vision task, and hope it is useful for future research. Moreover, since our method does not involve the sophisticated knowledge for modification of transformer block or additional input information, it is orthogonal and friendly to most of existing ViT designs and approaches.
	
	\noindent{\textbf{Our contributions.}}
	
	- We investigate the feasibility of leveraging recursive operation with sliced group self-attention in the vision transformers, which is a promising direction for establishing efficient transformers and has not been well-explored before. We conducted in-depth studies on the roles of recursion in transformers and conclude an effective scheme to use them for better parameter utilization.
	
	- We provide design principles, including the concrete format and comprehensive comparison to variants of SReT architectures, computational equivalency analysis, modified distillation, etc., in hope of enlightening future studies in compact transformer design and optimization.
	
	- We verify our method across a variety of scenarios, including vision transformer, all-MLP architecture of transformer variant, and neural machine translation (NMT) using transformers. Our model outperforms the state-of-the-art methods by a significant margin with fewer parameters.
	
	\section{Related Work}
	
	(i) \textbf{Transformer}~\cite{vaswani2017attention} was originally designed for natural language processing tasks and has been the dominant approach~\cite{devlin2019bert,yang2019xlnet,radford2019language,brown2020language,liu2019roberta} in this field. Recently, Vision Transformer (ViT)~\cite{dosovitskiy2021an} demonstrates that such multi-head self attention blocks can completely replace convolutions and achieve competitive performance on image classification. While it relied on pre-training on large amounts of data and transferring to downstream datasets. DeiT~\cite{touvron2020training} explored the training strategies and various data augmentation on ViT models, to train them on ImageNet-1K directly. Basically, DeiT can be regarded as a framework of ViT backbone + massive data augmentation + hyper-parameter tuning + hard distillation with tokens. After that, many extensions and variants of ViT models have emerged on image classification task, such as Bottleneck Transformer~\cite{srinivas2021bottleneck}, Multimodal Transformer~\cite{hendricks2021decoupling}, Tokens-to-Token Transformer~\cite{yuan2021tokens}, Spatial Pyramid Transformer~\cite{heo2021rethinking,wang2021pyramid}, Class-Attention Transformer~\cite{touvron2021going}, Transformer in Transformer~\cite{han2021transformer}, Convolution Transformer~\cite{wu2021cvt}, Shifted Windows Transformer~\cite{liu2021swin}, Co-Scale Conv-Attentional Transformer~\cite{xu2021co}, etc. 
	(ii) \textbf{Recursive operation} has been explored in NLP~\cite{liu2014recursive,dehghani2018universal,bai2018trellis,bai2019deep,bai2020multiscale,lan2019albert,pmlr-v139-chowdhury21a} and vision~\cite{liang2015recurrent,kim2016deeply,guo2019dynamic,liu2020cbnet} areas. In particular, DEQ~\cite{bai2019deep} proposed to find equilibrium points via root-finding in the weight-tied feedforward models like transformers and trellis for constant memory. UT~\cite{dehghani2018universal} presented the transformer with recurrent inductive bias of RNNs which is similar to our SReT format. However, these works ignored the complexity increased by recursive operation in designing networks. In this paper, we focus on utilizing recursion properly by approximating self-attention through multiple group self-attentions for building compact and efficient vision transformers.
	
	\section{Recursive Transformer} \label{main_method}
	
	\noindent{\textbf{Vanilla Transformer Block.}} A basic transformer block $\mathcal{F}$ consists of a Multi-head Self Attention (MHSA), Layer Normalization (LN), Feed-forward Network (FFN), and Residual Connections (RC). It can be formulated as:
	\begin{equation} \label{trans}
		\begin{aligned}\mathbf{z}_{\ell}^{\prime}=\text{MHSA}\left(\text{LN}\left(\mathbf{z}_{\ell-1}\right)\right)+\mathbf{z}_{\ell-1};          
			\mathbf{z}_{\ell}=\text{FFN}\left(\text{LN}\left(\mathbf{z}_{\ell}^{\prime}\right)\right)+\mathbf{z}_{\ell}^{\prime}; 
			i.e., \mathbf{z}_{\ell} = \mathcal{F}_{\ell-1}(\mathbf{z}_{\ell-1})
		\end{aligned}
	\end{equation}
	where $\mathbf{z}_{\ell}^{\prime}$ and $\mathbf{z}_{\ell-1}$ are the intermediate representations. $\mathcal{F}_{\ell}$ indicates the transformer block at $\ell$-th layer. $\ell \in \{0, 1,\dots, L\}$ is the layer index and $L$ is the number of hidden layers. The self-attention module is realized by the inner products with a scaling factor and a {\em softmax} operation, which is written as:
	\begin{equation}
		\operatorname{Attention}(Q, K, V)=\operatorname{Softmax}\left(Q K^{\top}/{\sqrt{d_{k}}}\right) V
	\end{equation}
	where $Q, K, V$ are {\em query}, {\em key} and {\em value} vectors, respectively. $1/\sqrt{d_k}$ is the scaling factor for normalization. Multi-head self attention further concatenates the parallel attention layers to increase the representation ability:\\ 
	$\text{MHSA}(Q, K, V)=\operatorname{Concat}\left(\operatorname{head}_{1}, \ldots,  \operatorname{ head }_{\mathrm{h}}\right) W^{O}$, 
	where $W^{O} \in \mathbb{R}^{h d_{v} \times d_{\text {model }}}$. $\text{head}_{\mathrm{i}}=\operatorname{Attention}\left(Q W_{i}^{Q}, K W_{i}^{K}, V W_{i}^{V}\right)$ are the projections with parameter matrices $W_{i}^{Q} \in \mathbb{R}^{d_{\text {model }} \times d_{k}}, W_{i}^{K} \in \mathbb{R}^{d_{\text {model }} \times d_{k}}, W_{i}^{V} \in \mathbb{R}^{d_{\text {model }} \times d_{v}}$. 
	The FFN contains two linear layers with a GELU non-linearity~\cite{hendrycks2016gaussian} in between
	\begin{equation}
		\mathrm{FFN}(x)= (\text{GELU}\left(\mathbf{z} W_{1}+b_{1}\right) ) W_{2}+b_{2}
	\end{equation}
	where $\mathbf{z}$ is the input. $W_{1},b_{1},W_{2},b_{2}$ are the two linear layers' weights and biases.
	
	\noindent{\textbf{Recursive Operation.}} In the original recursive module~\cite{sperduti1997supervised} for the language modality, the shared weights are recursively applied on a structured input which is among the complex inherent chains, so it is capable of learning deep structured knowledge. Recursive neural networks are made of architectural data and class, which is majorly proposed for model compositionality on NLP tasks. Here, we still use the sequence of patch tokens from the images as the inputs following the ViT model~\cite{dosovitskiy2021an}. And, there are no additional inputs used for feeding into each recursive loop of recursive block as used on structured data. Take two loops as an example for building the network, the recursive operation can be simplified:
	\begin{equation}
		\mathbf{z}_{\ell} = \mathcal{F}_{\ell-1}(\mathcal{F}_{\ell-1}(\mathbf{z}_{\ell-1}))
	\end{equation}
	The na\"ive recursive operation tends to learn a simple and trivial solution like the identity mapping by the optimizer, since the $\mathcal{F}_{\ell-1}$'s output and input are identical at the adjacent two depths (layers).
	
	\noindent{\textbf{Non-linear Projection Layer (NLL).}} NLL is placed between two recursive operations to enable the non-linear transformation between each block's output and input, to avoid learning trivial status for these recursive blocks by forcing nonequivalence on neighboring output and input. NLL can be formulated as:
	\begin{equation} \label{nlp}
		\text{NLL}(\mathbf{z}_{\ell-1})=\text{MLP}\left(\text{LN}\left(\mathbf{z}_{\ell-1}^{\prime}\right)\right)+\mathbf{z}_{\ell-1}^{\prime} 
	\end{equation}
	where MLP is a multi-layer projection as FFN, but has different {\em mlp ratio} for hidden features. We also use residual connection in it for better representation. As shown in Table~\ref{tab:increase-table}, more recursions will not improve accuracy without NLL.
	
	\noindent{\textbf{Recursive Transformer.}} A recursive transformer with two loops in every block is:
	\begin{equation}
		\mathbf{z}_{\ell} = \text{NLL}_2(\mathcal{F}_{\ell-1}(\text{NLL}_1(\mathcal{F}_{\ell-1}(\mathbf{z}_{\ell-1}))))
	\end{equation}
	where $\mathbf{z}_{\ell-1}$ and $\mathbf{z}_{\ell}$ are each recursive block's input and output. Different from MHSA and FFN that share parameters across all recursive operations within a block, $\text{NLL}_1$ and $\text{NLL}_2$ use the non-shared weights independently regardless of positioning within or outside the recursive blocks. 
	
	\noindent\textbf{Recursive All-MLP~\cite{tolstikhin2021mlpmixer} (an extension).} We can formulate it as:
	\begin{equation}
		\begin{aligned} \mathbf{U}_{*, i} &=\mathbf{X}_{*, i}+\mathbf{W}_{2} *\text{GELU}\left(\mathbf{W}_{1} *\text { LN }(\mathbf{X})_{*, i}\right),  \\ 
			\mathbf{Y}_{j, *} &=\mathbf{U}_{j, *}+\mathbf{W}_{4} *\text{GELU}\left(\mathbf{W}_{3} *\text { LN }(\mathbf{U})_{j, *}\right), \\ 
			\mathbf{Y}_{j, *} &=\mathcal{M}_{\ell-1}(\mathcal{M}_{\ell-1}(\mathbf{X}_{*, i}))
		\end{aligned}
	\end{equation}
	where the first and second lines are {\em token-mixing} and {\em channel-mixing} from~\cite{tolstikhin2021mlpmixer}. $\mathcal{M}_{\ell-1}$ is a MLP block, $C$ is the hidden dimension and $S$ is the number of non-overlapping image patches. NLL is not used here for simplicity.
	
	\noindent{\textbf{Gradients in A Recursive Block.}} Here, we simply use explicit backpropagation through the exact operations in the forward pass like gradient descent method since SReT has no constraint to obtain the equilibrium of input-output in recursions like DEQ~\cite{bai2019deep} and the number of loops can be small to control the network computation and depth. Our backward pass is more like UT~\cite{dehghani2018universal}. In general, the gradient of the parameters in each recursive block can be:
	\begin{equation}
		\begin{aligned} \frac{\partial \mathcal{L}}{\partial \mathbf{W}_\mathcal{F}} &=\frac{\partial \mathcal{L}}{\partial \mathbf{z}^{N}} \frac{\partial \mathbf{z}^{N}}{\partial \mathbf{W}_{\mathcal{F}}}+\frac{\partial \mathcal{L}}{\partial \mathbf{z}^{N}} \frac{\partial \mathbf{z}^{N}}{\partial \mathbf{z}^{N-1}} \frac{\partial \mathbf{z}^{N-1}}{\partial \mathbf{W}_{\mathcal{F}}}+\ldots\frac{\partial \mathcal{L}}{\partial \mathbf{z}^{N}} \frac{\partial \mathbf{z}^{N}}{\partial \mathbf{z}^{N-1}} \ldots \frac{\partial \mathbf{z}^{2}}{\partial \mathbf{z}^{1}} \frac{\partial \mathbf{z}^{1}}{\partial \mathbf{W}_{\mathcal{F}}} \\ &=\sum_{i=1}^{N} \frac{\partial \mathcal{L}}{\partial \mathbf{z}^{N}}\left(\prod_{j=i}^{N-1} \frac{\partial \mathbf{z}^{j+1}}{\partial \mathbf{z}^{j}}\right) \frac{\partial \mathbf{z}^{i}}{\partial \mathbf{W}_{\mathcal{F}}} \end{aligned}
	\end{equation}
	where $\mathbf{W}_\mathcal{F}$ is the parameters of recursive block. $\mathcal{L}$ is the objective function.
	
	\noindent{\textbf{Learnable Residual Connection (LRC) for Recursive Vision Transformers.}} He et al.~\cite{he2016identity} studied various strategies of shortcut connections on CNNs and found that the original residual design with pre-activation performs best. Here, we found simply adding learnable coefficients on each branch of residual connection can benefit to the performance of ViT following the similar discovery of literature~\cite{liu-etal-2019-self}. Formally, Eq.~\ref{trans} and Eq.~\ref{nlp} can be reformulated as:
	\begin{equation}
		\begin{aligned}
			\mathbf{z}_{\ell}^{\prime}&=\alpha*\text{MHSA}\left(\text{LN}\left(\mathbf{z}_{\ell-1}\right)\right)+\beta*\mathbf{z}_{\ell-1} ; \\ \mathbf{z}_{\ell}&=\gamma*\text{FFN}\left(\text{LN}\left(\mathbf{z}_{\ell}^{\prime}\right)\right)+ \delta*\mathbf{z}_{\ell}^{\prime} ; 
		\end{aligned}
	\end{equation}
	\begin{equation}
		\text{NLL}(\mathbf{z}_{\ell-1})=	\zeta*\text{MLP}\left(\text{LN}\left(\mathbf{z}_{\ell-1}^{\prime}\right)\right)+ \theta*\mathbf{z}_{\ell-1}^{\prime} 
	\end{equation}
	where $\alpha, \beta, \gamma, \delta, \zeta, \theta$ are the learnable coefficients. They are initialized as 1 and trained with other model's parameters simultaneously without restrictions. 
	
	\noindent{\textbf{Extremely Deep Transformers.}} Weight-sharing mechanism allows us to build a transformer with more than 100 layers still keeping a small model. We demonstrate empirically that the proposed method can significantly simplify the optimization when the transformer is scaled up to an exaggerated number of layers. 
	
	\begin{figure*}[t]
		\centering
		\includegraphics[width=0.96\textwidth]{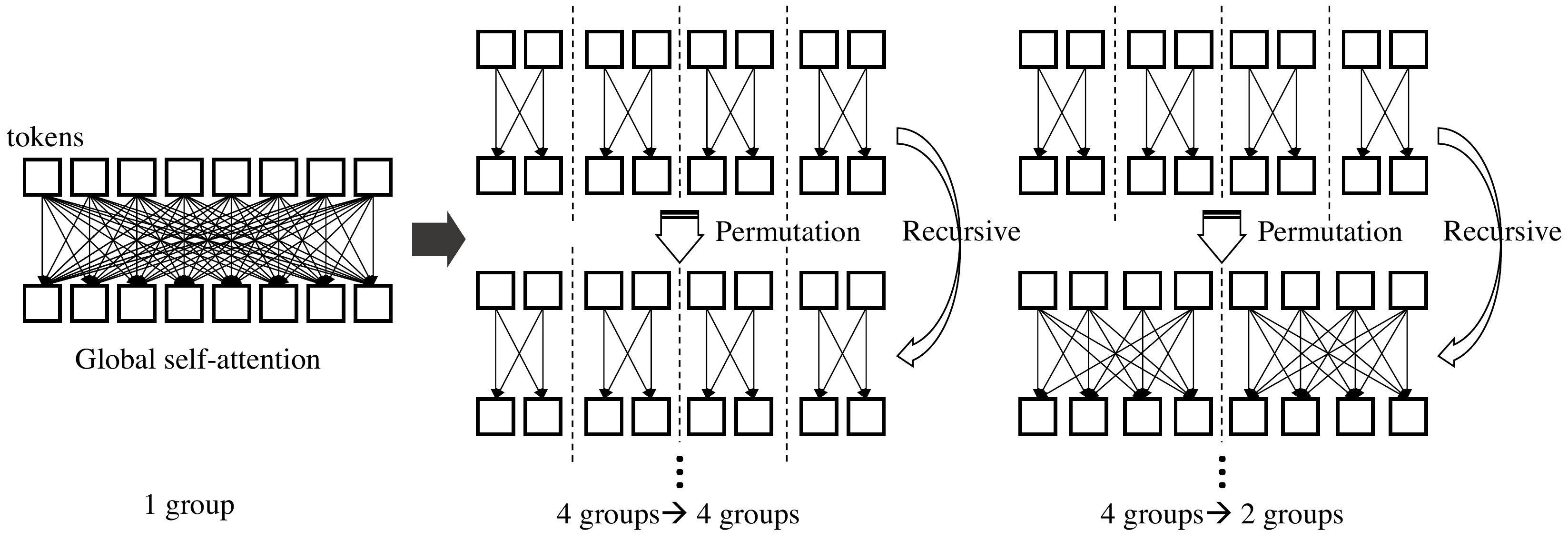} 
		\caption{Approximating global MHSA via sliced group MHSA with permutation.}
		\label{fig:pipeline123} 
	\end{figure*}
	
	\section{Approximating Global MHSA via Multi-Group MHSA} \label{approximating}
	
	Though recursive operation is adequate to provide better representation using the same number of parameters, the additional forward loop makes the overhead in training and inference increasing unnegligibly. To address the extra computational cost caused by recursion while maintaining the improved accuracy, we introduce an approximating method through multiple group self-attentions which is surprisingly effective in reducing FLOPs without compromising accuracy.
	
	\noindent{\textbf{Approximating Global Self-Attention in SReT.}} As shown in Fig.~\ref{fig:pipeline123}, a regular self-attention layer can be decoupled through multiple group self-attentions in a recursion manner with similar or even smaller computational cost. In general, the number of groups in different recursion can be the same or different depending on the requirements of FLOPs and accuracy trade-off. Such a strategy will not change the number of parameters while more groups can enjoy lower FLOPs but slightly inferior performance. We empirically verified that the decoupling scheme can achieve similar performance with significantly fewer FLOPs if using proper splitting of self-attention in a tolerable scope, as shown in Appendix.
	
	\noindent{\textbf{Computational Equivalency Analysis.}} In this subsection, we analyze the complexity of global ({i.e.,} original) and sliced group self-attentions and compare with different values of groups in a vision transformer.
	\begin{theorem} \label{Equivalency} 
		\ ({\em {\bf \em Equivalency of global and multiple group self-attentions with recursion on FLOPs.}) {\em Let $\{\textbf{N}_{\ell}, \textbf{G}_{\ell}\}\in \mathbb{R}^1$, when  $\textbf{N}_{\ell}\!=\!\textbf{G}_{\ell}$,  $\textbf{FLOPs}$(1 $\textbf{V-SA}$)= $\textbf{FLOPs}$($\textbf{N}_{\ell}\times\!$ Recursion with $\textbf{G}_{\ell}\times \textbf{G-SAs}$)}.}
		The complexity $\textbf{C}$ of global and group self-attentions can be calculated as: (For simplicity, here we assume \#groups and vector dimensions in each recursive operation are the same.)
		\begin{equation} \label{trans1234}
			{\bm {\textbf{C}_\text{G-SA}} = \frac{\textbf{N}_{\ell}}{\textbf{G}_{\ell}}\times \bm {\textbf{C}_\text{V-SA} }} 
		\end{equation}
		where $\textbf{N}_{\ell}$ and $\textbf{G}_{\ell}$ are the numbers of recursion and group MHSA in layer $\ell$, i.e., $\ell$-th recursive block. $\textbf{V-SA}$ and $\textbf{G-SA}$ represent the vanilla and group MHSA.
	\end{theorem}
	The proof is provided in Appendix. The insight provided by Theorem~\ref{Equivalency} is at the core of our method to control the complexity and its various benefits on better representations. Importantly, the computation of self-attention through the ``slice'' paralleling is equal to the vanilla self-attention. We can observe that when {\em ${\textbf{N}_\ell= \textbf{G}_\ell}$}, {\em ${\textbf{C}_\text{V-SA} \approx  \textbf{C}_\text{G-SA}}$}\footnote{In practice, the FLOPs of the two forms are not identical as self-attention module includes extra operations like softmax, multiplication with scale and attention values, which will be multiples by the recursive operation.} and if {\em ${\textbf{N}_\ell\!<\! \textbf{G}_\ell}$}, {\em $ \bm{\bm \textbf{C}_\text{G-SA}\!<\!\textbf{C}_\text{V-SA}}$}, we can use this property to reduce the FLOPs in designing ViT. 
	
	\noindent{\bf Empirical observation:}  {\em When $\textbf{FLOPs}$(recursion $\!$ + $\!\textbf{G-SA}$) $\!\approx\!$ $\textbf{FLOPs}$($\textbf{V-SA}$), $\textbf{Acc.}$(recursion + $\textbf{G-SAs}$) $>$ $\textbf{Acc.}$($\textbf{V-SA}$).}
	
	\begin{table}[t]
		\centering
		\caption{Representation ability with global/group self-attentions.}
		\label{tab:rep} \vspace{0.08in}
		\resizebox{.85\textwidth}{!}{
			\begin{tabular}{l|c|c|c}
				\toprule[1.1pt]
				Method         & \#Params (M) & FLOPs (B) & Top-1 Acc. (\%) \\ \hline
				Baseline (PiT~\cite{heo2021rethinking})   &    4.9 & 0.7  &  73.0  \\ \hline
				SReT (global self-attention w/o loop)         &  \bf 4.0 & 0.7 & 73.6 \\ 
				SReT (group self-attentions w/ loops)  &  \bf 4.0 & 0.7 &  \bf 74.0 \\ 
				\bottomrule[1.1pt] 
		\end{tabular}}
	\end{table}
	
	We employ ex-tiny model to evaluate the performance of global self-attention and sliced group self-attention with recursion. As shown in Table~\ref{tab:rep}, we empirically verify that, with the similar computation, group self-attention with recursion can obtain better accuracy than vanilla self-attention.

	\noindent{\textbf{Analysis: Where is the benefit from in SReT?}} Theoretical analysis on recursion could further help understand the advantage behind, while it is difficult and prior literature on this always proves it empirically. Here, we provide some basic theoretical explanations from the optimization angle for better understanding this approach. One is the enhanced gradients accumulation. Let $g_{t}$=$\nabla_{\theta} f_{t}(\theta)$ denote the gradient, we take Adam optimizer~\cite{kingma2014adam} as an example, na\"{i}ve parameter update is $\theta_{t} \!\leftarrow\! \theta_{t-1}\!-\!\alpha \cdot \widehat{m}_{t} /\left(\sqrt{\widehat{v}_{t}}+\epsilon\right)$ where the gradients {\em w.r.t.} stochastic objective at timestep $t$ is $g_{t} \!\leftarrow\! \nabla_{\theta} f_{t}\left(\theta_{t-1}\right)$, here we omit first and second moment estimate formulae. After involving recursion (here $\textbf{NLL}$ guarantees $\widehat{m}^i_{t}$, ${\widehat{v}^i_{t}}$'s discrepancy), the new updating is: $\theta_{t} \!\leftarrow\! \theta_{t-1}\!-\!\sum_{i=1}^{\bm{N}}\alpha \cdot \widehat{m}^i_{t} /\left(\sqrt{\widehat{v}^i_{t}}+\epsilon\right)$ where $\bm N$ is the number of recursion loops. Basically, recursion enables more updating/tuning of parameters in the same iteration, so that the learned weights are more aligned to the loss function, and the performance is naturally better.
	
	\section{Experiments}
	
	In this section, we first empirically verify the proposed SReT on image classification task with self-attention~\cite{vaswani2017attention} and all-MLP~\cite{tolstikhin2021mlpmixer} architectures, respectively. We also perform detailed ablation studies to explore the optimal hyper-parameters of our proposed network. Then, we extend it to the neural machine translation (NMT) task to further verify the generalization ability of the proposed approach. Finally, we visualize the evolution of learned coefficients in LRC and intermediate activation maps to better understand the behaviors and properties of our proposed model. Our experiments are conducted on CIAI cluster.
	
	\subsection{Datasets and Experimental Settings} \label{dataset_setting}
	(i) {\bf ImageNet-1K}~\cite{deng2009imagenet}: ImageNet-1K is a standard image classification dataset, which contains 1K classes with a total number of 1.2 million training images and 50K validation images. Our models are trained on this dataset solely without additional images; (ii) {\bf IWSLT'14 German to English (De-En)} dataset~\cite{IWSLT14}: It contains about 160K sentence pairs as the training set. We train and evaluate models following the protocol~\cite{IWSLT14_fair}; (iii) {\bf WMT'14 English to German (En-De)} dataset~\cite{WMT14}: The WMT’14 training data consists of 4.5M sentences pairs (116M English words, 110M German words). We use the same setup as~\cite{luong2015effective}.
	
	\noindent{\bf{Settings:}} Our detailed training settings and hyper-parameters are shown in Appendix.  On ImageNet-1K, our backbone network is a spatial pyramid~\cite{heo2021rethinking} architecture with stem structure following~\cite{shen2017dsod}.
	
	\noindent{\textbf{Soft distillation strategy.}} On vision transformer, DeiT~\cite{touvron2020training} proposed to distill tokens together with hard predictions from the teacher. They stated that using one-hot label with hard distillation can achieve the best accuracy. This seems counterintuitive since soft labels can provide more subtle differences and fine-grained information of the input. In this work, through a proper distillation design, our soft label based distillation framework (one-hot label is not used) consistently obtained better performance than DeiT\footnote{We observed a minor 
		issue of soft distillation implementation in DeiT (\url{https://github.com/facebookresearch/deit/blob/main/losses.py\#L56}). Basically, it is unnecessary to use {\em logarithm} for teacher's output (logits) according to the formulation of KL-divergence or cross-entropy. Adding {\em log} on both teacher and student's logits will make the results of KL to be extremely small and intrinsically negligible. 
		We argue that soft labels can provide fine-grained information for distillation, and consistently achieve better results using soft labels in a proper way than {\em one-hot label + hard distillation}, as shown in Sec.~\ref{ablation_study}.}. Our loss is a soft version of cross-entropy between teacher and student's outputs as used in~\cite{shen2021is,romero2014fitnets,shen2019meal,bagherinezhad2018label}:
	$
	{{\mathcal{L}}_{CE}}({\mathcal{S}_\mathbf{W} }) =  - \frac{1}{N}{\sum_{i = 1}^N {{\bf P}_{{\mathcal{T}_\mathbf{W} }}({\mathbf{z}})\log } } {\bf P}_{{\mathcal{S}_\mathbf{W} }}({\mathbf{z}}),
	$
	where ${\bf P}_{\mathcal{T}_\mathbf{W}}$ and ${\bf P}_{\mathcal{S}_\mathbf{W}}$ are the outputs of teacher and student, respectively. More details can be referred to Appendix.
	
	\subsection{Na\"ive Recursion on Transformer}
	
	\begin{wrapfigure}{r}{4.2cm}
		\vspace{-0.4in}\centering
		\includegraphics[width=0.95\linewidth]{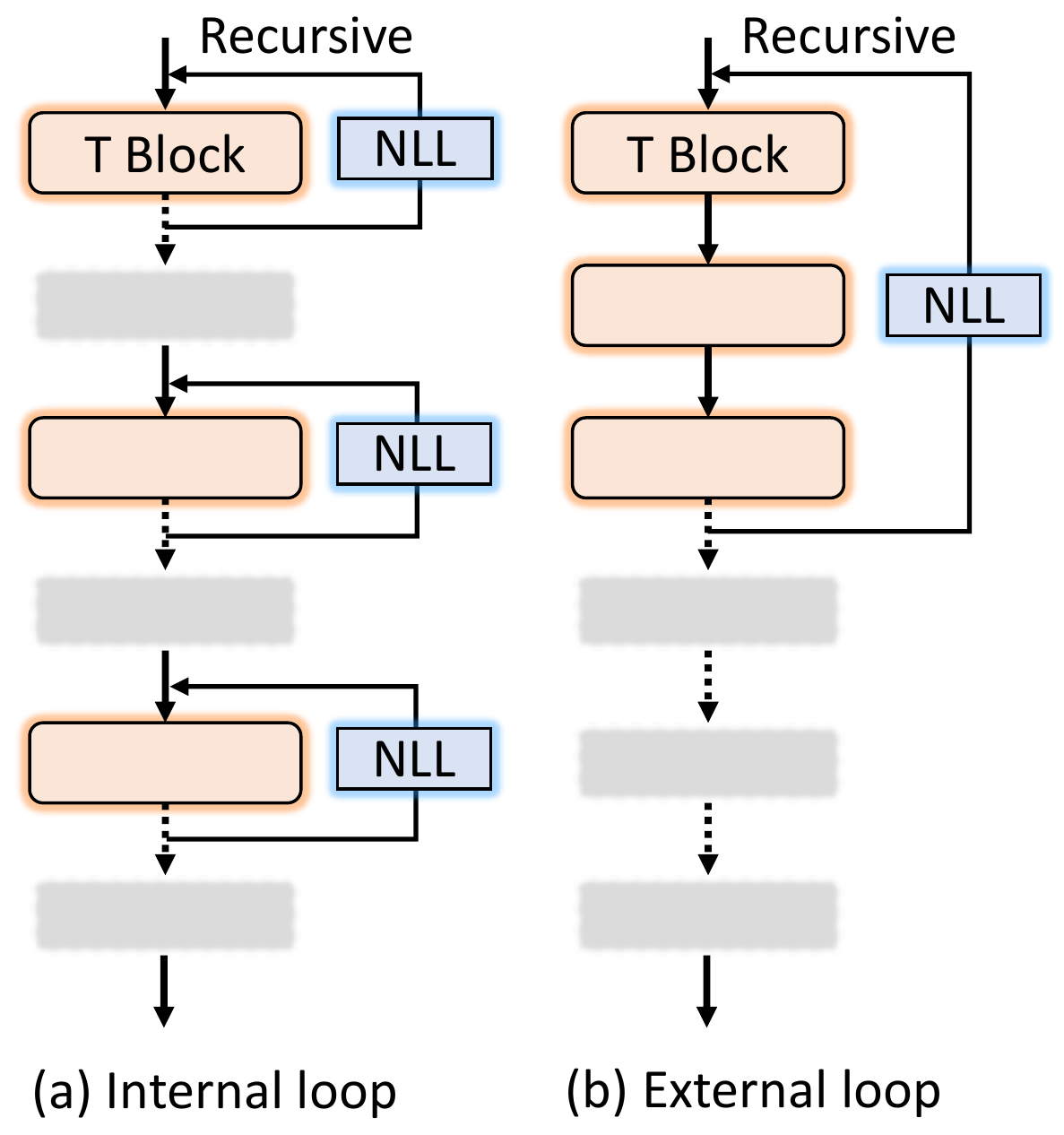}\vspace{-0.1in}
		\caption{Paradigms of recursive designs in transformer.}
		\label{fig:inside_outside}\vspace{-0.3in}
	\end{wrapfigure}
	
	In this section, we examine the effectiveness of proposed recursion using DeiT training strategies. We verify the following two fashions of recursion. 
	
	\noindent{\bf{Internal and external loops.}} As illustrated in Fig.~\ref{fig:inside_outside}, there are two possible recursion designs on transformer networks. One is the internal loop that repeats every block separately. Another one is the external loop that cyclically executes all blocks together. Although external loop design can force the model being more compact as it shares parameters across all blocks with fewer non-shared NLL layers, we found such a structure is inflexible with limited representation ability. We conducted a comparison with 12 layers of basic transformers with 2$\times$ recursive operation and the results are: external 67.0\% (3.2M) {\em vs.} internal 67.6\% (3.0M) $|$ 70.3\% (3.9M). In the following experiments, we use the internal recursive design as our default setting.
	
	\begin{table}[t]
		\begin{minipage}[c]{0.70\textwidth}
			\begin{flushright}
			{\renewcommand{\arraystretch}{1.15}
				\resizebox{0.9\textwidth}{!}{
					\begin{tabular}{l|c|c}
						\toprule
						& \#Params (M) & Top-1 (\%) \\ \hline
						Baseline                      & 5.7       & 72.2       \\ \hline
						Recursion \textcolor{gray}{$_\text{w/o NLL}$}   & 3.8       & 72.5       \\ 
						Recursion + NLL          & 5.0       & 74.7       \\ 
						Recursion + NLL - Class Token & 5.0       & 75.0       \\ 
						Recursion + NLL + LRC               & 5.0       & 75.2       \\
						Recursion + NLL + Stem              & 5.0       & 76.0       \\  \hline
						Recursion  (Full Components)             & 5.0       & \bf 76.8       \\ \hline\hline
						GT+Hard Distill~\cite{touvron2020training}          & 5.0       & 77.5       \\ 
						Soft Distill (Ours)             & 5.0       & \bf 77.9       \\ 
						\bottomrule
			\end{tabular}}}
		\end{flushright}
		\end{minipage}\hfill 
		\begin{minipage}[c]{0.28\textwidth}\vspace{-0.1in}
			\begin{flushleft}
				\caption{Effectiveness of various designs on ImageNet-1K val set. Please refer to Sec.~\ref{ablation_study} and our Appendix for more details. In this ablation study, the backbone is \texttt{SReT-TL} model using spatial pyramid architecture.} 
				\label{tab:my-table-ablation}
			\end{flushleft}
		\end{minipage}
	\end{table}
	
	\subsection{Ablation Studies} \label{ablation_study}
	
	The overview of our ablation studies is shown in Table~\ref{tab:my-table-ablation}. The first row presents the baseline, the second group is the different structures indicated by the used factors. The last is the comparison of KD. We also verify the following designs.
	
\begin{figure*}[b]
		\centering \vspace{-0.2in}
		\includegraphics[width=0.98\linewidth]{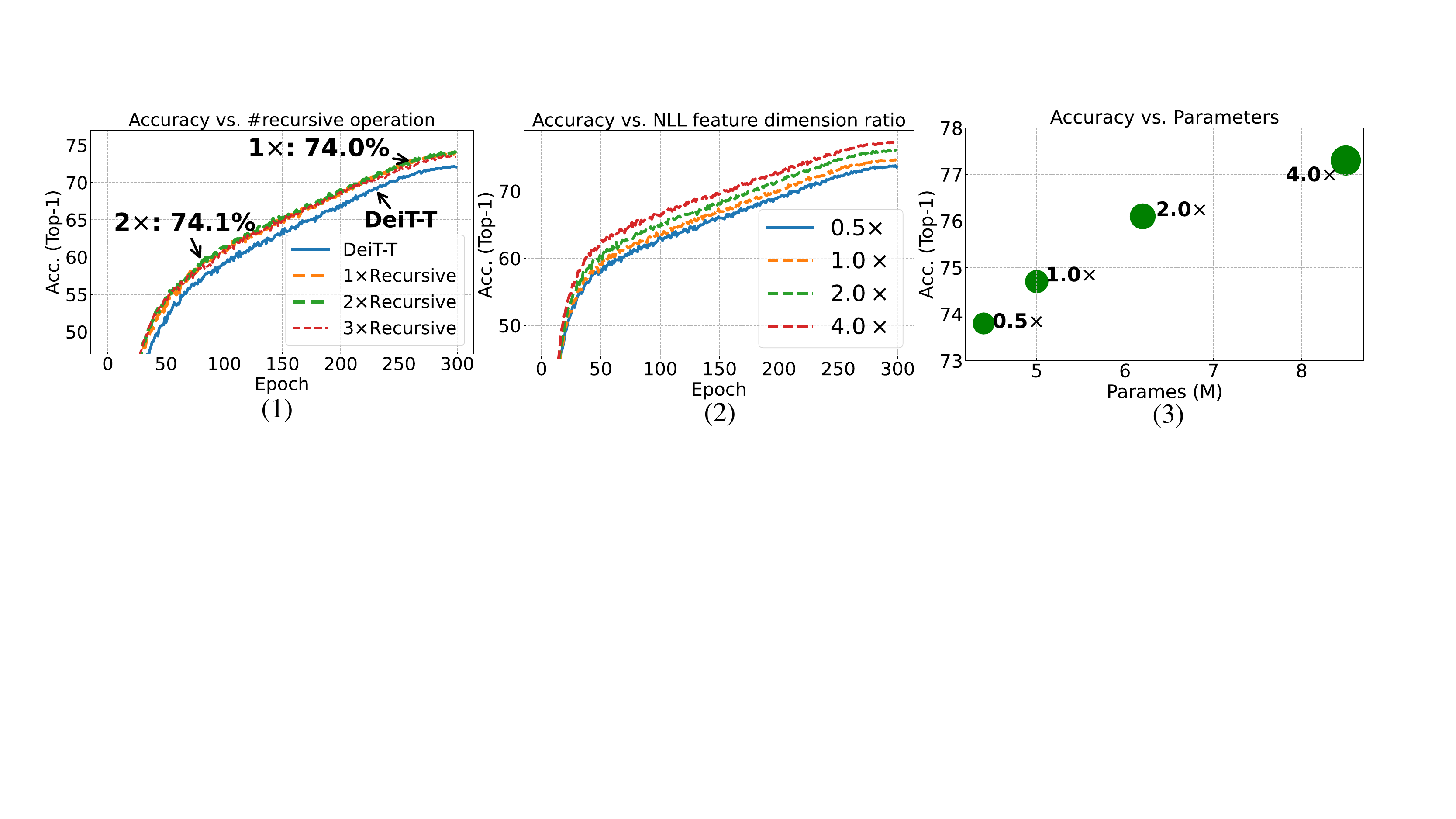}\vspace{-0.1in}
		\caption{A comprehensive ablation study on different design factors.}
		\label{fig:ablation_all_1}
	\end{figure*}

	\noindent{\textbf{Architecture configuration.}} As in Table~\ref{tab:my-table_sota}, \texttt{SReT-T} is our tiny model which has {\em mlp ratio} = 3.6 in FFN and 4.0 for \texttt{SReT-TL}. More details about these architectures are provided in our Appendix. To examine the effectiveness of recursive operation, we conduct different loops of na\"ive recursion on DeiT-T. The results of accuracy curves on validation data are shown in Fig.~\ref{fig:ablation_all_1} (1), we can see $2\times$ is slightly better than $1\times$ and the further boost is marginal, while the $1\times$ is much faster for executing. Thus, we use this in the following experiments.
	
	\noindent{\textbf{NLL configuration.}} NLL is a crucial factor for size and performance since the weights in it are not shared. To find an optimal trade-off between model compactness and accuracy, we explore the NLL ratios in Fig.~\ref{fig:ablation_all_1} (2, 3). Generally, a larger NLL ratio can achieve better performance but the model size increases accordingly. We use 1.0 in our \texttt{SReT-T} and \texttt{SReT-TL}, and 2.0 in our \texttt{SReT-S}.
	
	\noindent{\textbf{Different permutation designs and groups numbers.}} We explore the different permutation designs and the principle of choosing group numbers for better accuracy-FLOPs trade-off. We propose to insert permutation and inverse permutation layers to preserve the input's order information after the sliced group self-attention operation. The detailed formulation of this module, together with recursions and their result analyses are given in our Appendix.
	
	\noindent{\textbf{Distillation.}}  To examine the effectiveness of our proposed soft distillation method, we conduct the comparison of {\em one-hot label + hard distillation} and {\em soft distillation only}. The backbone network is \texttt{SReT-T}, all hyper-parameters are the same except the loss functions. The accuracy curves are shown in our Appendix. Our result 77.7\% is significantly better than the baseline 77.1\%.
	
	\noindent{\textbf{Throughput evaluation.}} In Table~\ref{tab:my-table2}, we provide the throughput comparisons with DeiT and Swin on one NVIDIA GeForce RTX 3090 which can directly reflect the real inference speed and time consumption. We highlight that our method obtains significantly fewer params and FLOPs with better throughput.
	
	\begin{table}[t]
		\centering
		\caption{Throughput evaluation of {SReT} and baselines.}
		\label{tab:my-table2} \vspace{0.08in}
		\resizebox{0.95\textwidth}{!}{
			\begin{tabular}{l|l|l|l|l}
				\hline
				DeiT-T        & FLOPs: 1.3B  & \#Params: 5.7M & Acc.: 72.2\% & Throughput: 3283.49 img/s \\ 
				\bf SReT-ExT        & FLOPs: \bf 0.7B{\color{black}$^{\bf \downarrow46.2\%}$}  & \#Params: \bf 4.0M{\color{black}$^{\bf \downarrow29.8\%}$} & Acc.: \bf 74.0\%{\color{black}$^{\bf \uparrow1.8\%}$} & Throughput: \bf 3473.43 img/s \\  
				Swin-T        & FLOPs: 4.5B  & \#Params: 29.0M & Acc.:  81.3\% & Throughput: 1071.43 img/s \\
				\bf SReT-S         & FLOPs: \bf 4.2B{\color{black}$^{\bf \downarrow6.7\%}$}  & \#Params: \bf 20.9M{\color{black}$^{\bf \downarrow27.9\%}$} & Acc.: \bf 81.9\%{\color{black}$^{\bf \uparrow0.6\%}$} & Throughput: \bf 1101.84 img/s \\ \hline
		\end{tabular}}
	\end{table}
	
	\begin{table*}[t]
		\centering
		\caption{Comparison of Top-1 (\%) on ImageNet-1K with state-of-the-art methods. $\divideontimes$ denotes the model is trained without the proposed group self-attention approximation. {\em Fine-tuning on large resolution} is highlighted by gray color.}
		\label{tab:my-table_sota} \vspace{0.08in}
		{\renewcommand{\arraystretch}{1.0}
			\resizebox{0.85\textwidth}{!}{%
				\begin{tabular}{l|c|c|c|c}
					\toprule
					\bf Method & \bf Resolution & \bf \#Params (M) & \bf FLOPs (B) & \bf Top-1 (\%) \\ \hline
					DeiT-T~\cite{touvron2020training}  & 224   &   5.7  &    1.3   &   72.2      \\ 
					PiT-T~\cite{heo2021rethinking} &  224    &  4.9   &   0.7   &   73.0       \\ 
					\bf SReT-ExT (Ours) &  224   &  \bf 4.0 &     0.7   &  \bf  74.0    \\ \hline
					DeiT-T~\cite{touvron2020training}  & 224   &   5.7  &    1.3   &   72.2      \\ 
					\bf SReT-$\divideontimes$T (Ours) &  224   &  \bf ~~~~~~~4.8$^{\downarrow \bf 15.8\%}$  &     1.4   &  \bf 76.1     \\ 
					\bf SReT-T (Ours) &  224   &  \bf 4.8  &  ~~~~~~~~\bf 1.1$^{\downarrow \bf 21.4\%}$   &  \bf   76.0   \\ \hline
					DeiT-T$_{ Distill}$~\cite{touvron2020training}  & 224   &   5.7  &     1.3   &   74.5        \\ 
					\bf SReT-$\divideontimes$T$_{Distill}$ (Ours) &  224   &  \bf 4.8  &   1.4    &  \bf 77.7     \\ 
					\bf SReT-T$_{ Distill}$ (Ours) &  224   & \bf  4.8  &   \bf ~~~~~~~1.1$^{\downarrow \bf 21.4\%}$   &   \bf 77.6   \\ 
					\rowcolor{mygraylite}  \bf SReT-$\divideontimes$T$_{ Distill\&384\uparrow}$ (Ours) &  384   &   4.9  &  6.4     & 79.7   \\ 
					\rowcolor{mygraylite}  \bf SReT-T$_{ Distill\&384\uparrow}$  (Ours) &  384   &   4.9  &  \bf ~~~~~~~4.3$^{\downarrow \bf 32.8\%}$    &  79.6  \\ \hline \hline
					DeiT-T~\cite{touvron2020training}  & 224   &   5.7  &    1.3   &   72.2      \\ 
					AutoFormer-Tiny~\cite{AutoFormer} &  224   &   5.7  &     1.3   &  74.7     \\ 
					CoaT-Lite Tiny~\cite{xu2021co} &  224    &  5.7   &    1.6    &   76.6     \\ 
					\bf SReT-$\divideontimes$TL (Ours)  & 224   &  \bf ~~~~~~~5.0$^{\downarrow \bf 12.3\%}$  &   1.4    &   \bf 76.8   \\ 
					\bf SReT-TL (Ours)  & 224   & \bf  5.0  &  \bf ~~~~~~~1.2$^{\downarrow \bf 14.3\%}$    &   \bf 76.7   \\ \hline
					\bf SReT-$\divideontimes$TL$_{ Distill}$ (Ours) &  224    &  5.0   &   1.4    &   77.9    \\ 
					\bf SReT-TL$_{ Distill}$ (Ours) &  224    &  5.0   &   1.2    &   77.7    \\ 
					\rowcolor{mygraylite} \bf SReT-$\divideontimes$TL$_{ Distill\&384\uparrow}$ (Ours) &  384   &   5.1  &    6.6    &  80.0     \\ 
					\rowcolor{mygraylite} \bf SReT-TL$_{ Distill\&384\uparrow}$ (Ours) &  384   &   5.1  &   \bf ~~~~~~~4.4$^{\downarrow \bf 33.3\%}$    &   79.8    \\ \hline \hline
					ViT-B/16~\cite{dosovitskiy2021an} & 384   &   86.0  &  55.4    &   77.9   \\ 
					DeiT-S~\cite{touvron2020training}  & 224   &   22.1  &  4.6     &   79.8    \\ 
					PVT-S~\cite{wang2021pyramid} & 224    & 24.5 &    3.8   &   79.8   \\ 
					PiT-S~\cite{heo2021rethinking} &  224    &  23.5   &   2.9    &   80.9      \\ 
					T2T-ViT$_t$-14~\cite{yuan2021tokens} &  224    &  21.5  & 5.2  &  80.7    \\ 
					TNT-S~\cite{han2021transformer} &  224    &  23.8   &   5.2    &   81.3    \\ 
					Swin-T~\cite{liu2021swin} &  224    &  29.0   &   4.5    &   81.3    \\ 
					\bf SReT-$\divideontimes$S (Ours) &  224   & \bf   ~~~~~~~20.9$^{\downarrow \bf 27.9\%}$  &   4.7    & \bf 82.0    \\ 
					\bf SReT-S (Ours) &  224   & \bf  20.9  &   \bf ~~~~~~~4.2$^{\downarrow \bf 10.6\%}$   & \bf  81.9   \\ \hline
					PiT-S$_{ Distill}$~\cite{heo2021rethinking} &  224    &  23.5   &   2.9   &   81.9    \\ 
					DeiT-S$_{ Distill}$~\cite{touvron2020training}  & 224   &   22.1  &    4.6   &   81.2    \\ 
					T2T-ViT$_t$-14$_{ Distill}$~\cite{yuan2021tokens} &  224    &  21.5   &  5.2     &   81.7      \\ 
					\bf SReT-$\divideontimes$S$_{ Distill}$ (Ours) &  224   &  \bf 20.9  &     4.7   & \bf 82.8    \\ 
					\bf SReT-S$_{ Distill}$ (Ours) &  224   &  \bf 20.9  &    \bf ~~~~~~~4.2$^{\downarrow \bf 10.6\%}$     & \bf  82.7   \\ \hline
					\rowcolor{mygraylite} \bf SReT-$\divideontimes$S$_{ Distill\&384\uparrow}$ (Ours) &  384   &  \bf 21.0  &   18.5     & \bf   83.8  \\ 
					\rowcolor{mygray} \bf SReT-$\divideontimes$S$_{ Distill\&512\uparrow}$ (Ours) &  512   &  \bf 21.3  &   42.8  & \bf  84.3   \\ 
					\bottomrule
		\end{tabular}}}  \vspace{-0.08in}
	\end{table*}
	
	\begin{figure}[b]
	    \centering \vspace{-0.2in}
		\includegraphics[width=0.85\linewidth]{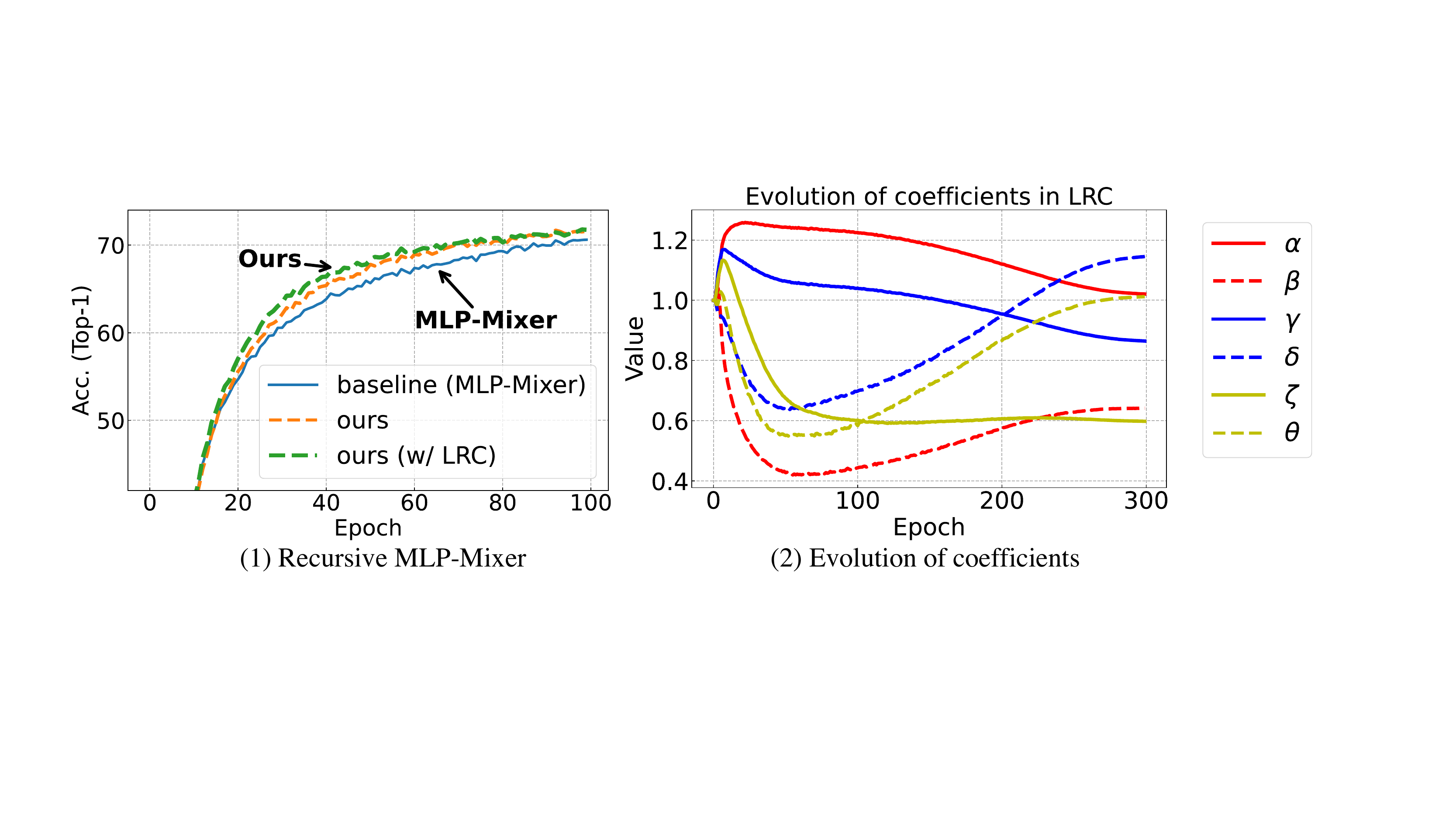}
		\caption{(1) ImageNet-1K results on All-MLP. (2) Evolution of coefficients.}
		\label{fig:mix_coefficients}
	\end{figure}
	
	\begin{figure*}[t]
		\centering
		\includegraphics[width=0.99\linewidth]{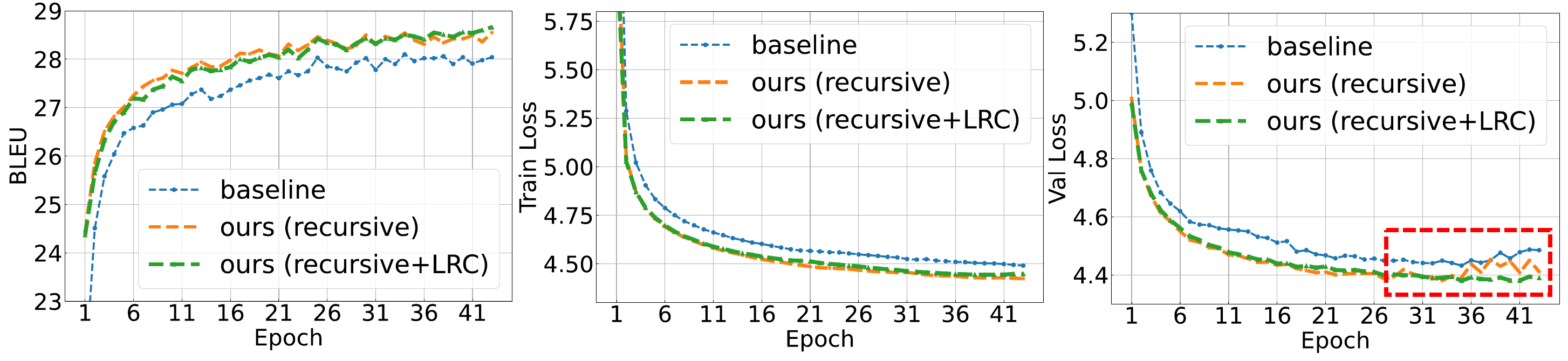}
		\caption{Comparison of BLEU, training loss and val loss on WMT14 En-De.}
		\label{fig:NMT_results}
	\end{figure*}
	
	\subsection{Comparison with State-of-the-art Approaches}
	
	A summary of our main results is shown in Table~\ref{tab:my-table_sota}, our \texttt{SReT-ExT} is better than PiT-T by 1.0\% with 18.4\%$\downarrow$ parameters. \texttt{SReT-T} also outperforms DeiT-T by 3.8\% with 15.8\%$\downarrow$ parameters and 15.4\%$\downarrow$ FLOPs. Distillation can help improve the accuracy by 1.6\% and fine-tuning on large resolution further boosts to 79.6\%. Moreover, our \texttt{SReT-S} is consistently better than state-of-the-art Swin-T, T2T, etc., on accuracy, model size and FLOPs, which demonstrates the superiority and potential of our architectures in practice.
	
	\subsection{All-MLP Architecture}\label{all_mlp_a}
	
	\noindent{\textbf{MLP-Mixer}}~\cite{tolstikhin2021mlpmixer} (Baseline), MLP-Mixer+Recursion and MLP-Mixer+Recursion +LRC: Mixer is a recently proposed plain design that is based entirely on multi-layer perceptrons (MLPs). We apply our recursive operation and LRC on MLP-Mixer to verify their generalization. Results are shown in Fig.~\ref{fig:mix_coefficients} (1), our method is consistently better than the baseline using the same training protocol.
	
	\begin{figure}[h]
		\centering
		\includegraphics[width=0.72\textwidth]{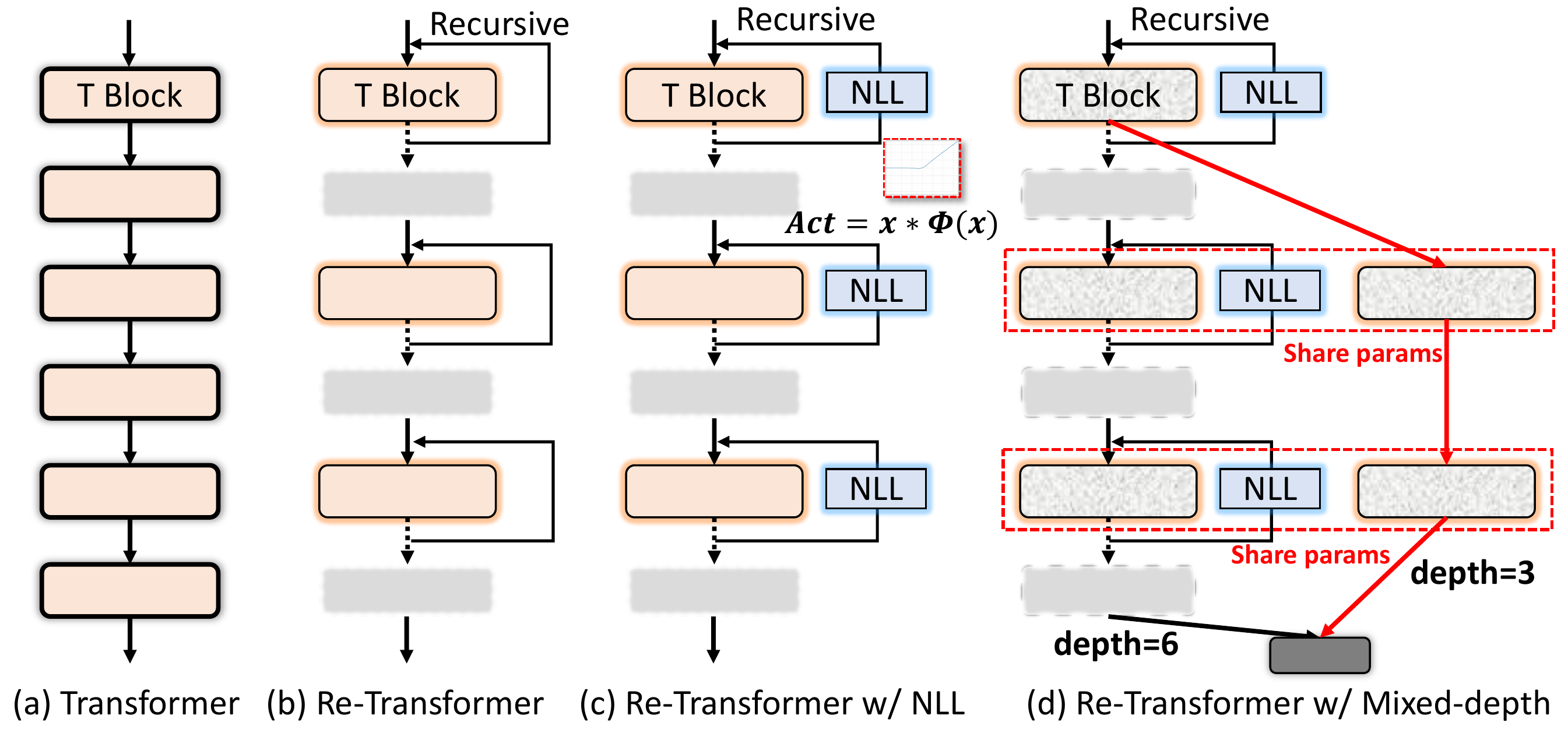} 
		\caption{Illustration of recursive transformer with different designs.}
		\label{fig:pipeline} 
	\end{figure}
	
	\begin{figure}[b]
        \centering \vspace{-0.2in}
		\includegraphics[width=0.87\linewidth]{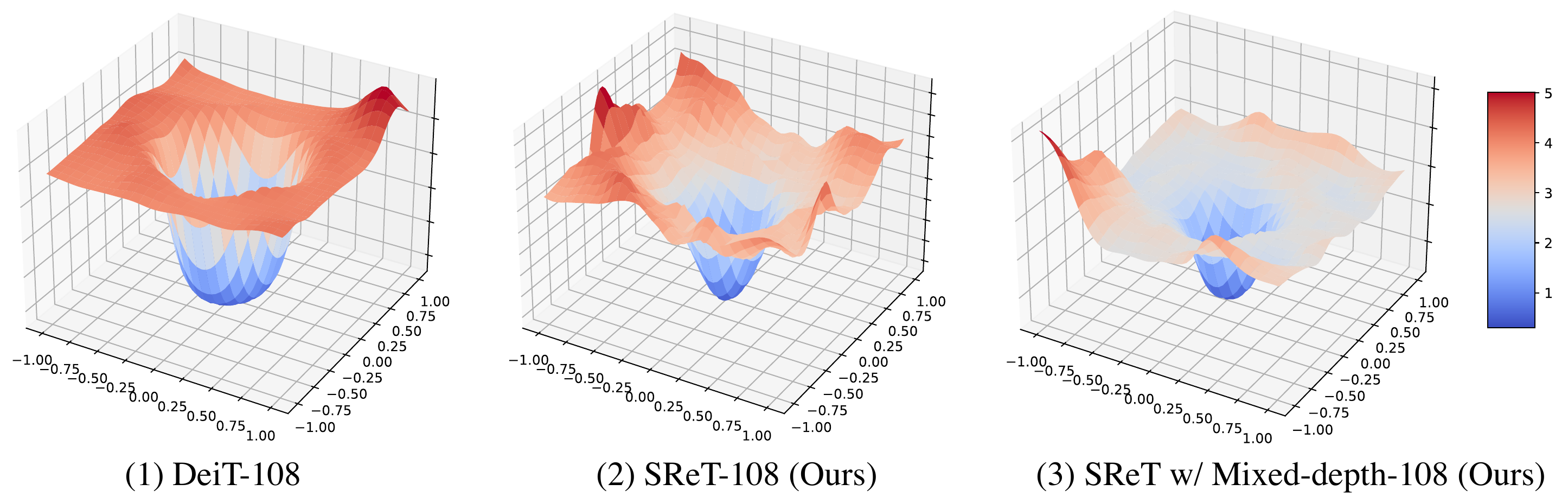}\vspace{-0.12in}
		\caption{The actual optimization landscape from \texttt{DeiT-108}, our \texttt{SReT-108} and \texttt{SReT-108 mixed-depth} models.}
		\label{fig:landscape_actual}
	\end{figure}
	
	\subsection{Neural Machine Translation}\label{machine_translation}
	
	In this section, we compare the BLEU scores~\cite{papineni-etal-2002-bleu} of vanilla transformer~\cite{vaswani2017attention} and ours on the WMT14 En-De and IWSLT'14 De-En (Appendix) using fairseq toolkit~\cite{fair_sque}. IWSLT'14 De-En is a relatively small dataset so the improvement is not as significant as on WMT14 En-De. The results are shown in Fig.~\ref{fig:NMT_results}, we can see our method is favorably better than the baseline. Without LRC, the model slightly converges faster, but the final accuracy is inferior to using LRC. Also, LRC makes the training process more stable, as shown in the red dashed box.
	
	\subsection{Landscape Visualizations of DeiT and Our Mixed-depth SReT} \label{landscape}
	\noindent{\textbf{Explicit mixed-depth training.}} The recursive neural network enables to train the model in a mixed-depth scheme. As shown in Fig.~\ref{fig:pipeline} (d), the left branch is the subnetwork containing recursive blocks, while the right is the blocks without sharing the weights on depth, but their weights are re-used with the left branch. In this structure, the two branches take inputs from the same stem block. Mixed-depth training offers simplified optimization by performing operations parallelly and prevents under-optimizing when the network is extremely deep.
	
	\noindent{\textbf{Benefits of mixed-depth training.}} The spin-off benefit of sliced recursion is the feasibility of mixed-depth training, which essentially is an {\em explicit deep supervision} scheme as the shallow branch receives stronger supervision that is closer to the final loss layer, meanwhile, weights are shared with the deep branch. 
	
	Inspired by~\cite{li2018visualizing}, we visualize the landscape of baseline DeiT-108 and our \texttt{SReT-108} \& \texttt{SReT-108 mixed-depth} models to examine and analyze the difficulty of optimization on these three architectures. The results are illustrated in Fig.~\ref{fig:landscape_actual}, we can observe that DeiT-108 is more chaotic and harder for optimization with a deeper local minimum than our mixed-depth network. This verifies the advantage of our proposed network structure for simpler optimization.
	
	\subsection{Analysis and Understanding}\label{ana_und}
	
	Here, we provide two visualizations regarding LRC and learned response maps. \\
	\noindent{\textbf{Evolution of LRC coefficients.}} As shown in Fig.~\ref{fig:mix_coefficients} (2), we plot the evolution of learned coefficients in the first block. We can observe that the coefficients on the identity mapping ($\alpha, \gamma, \zeta$) first go up and then down as the training continues. This phenomenon indicates that, at the beginning of model training, the identify mapping plays a major role in the representations. After $\sim$50 epochs of training, the main branch is becoming increasingly important. Once the training is complete, in FFN and NLL, the main branch exceeds the residual connection branch while on MHSA it is the opposite. We believe this phenomenon can inspire us to design a more reasonable residual connection structure in ViT.
	
\begin{figure}[t]
	\begin{minipage}[c]{0.6\textwidth}
			\includegraphics[width=0.99\linewidth]{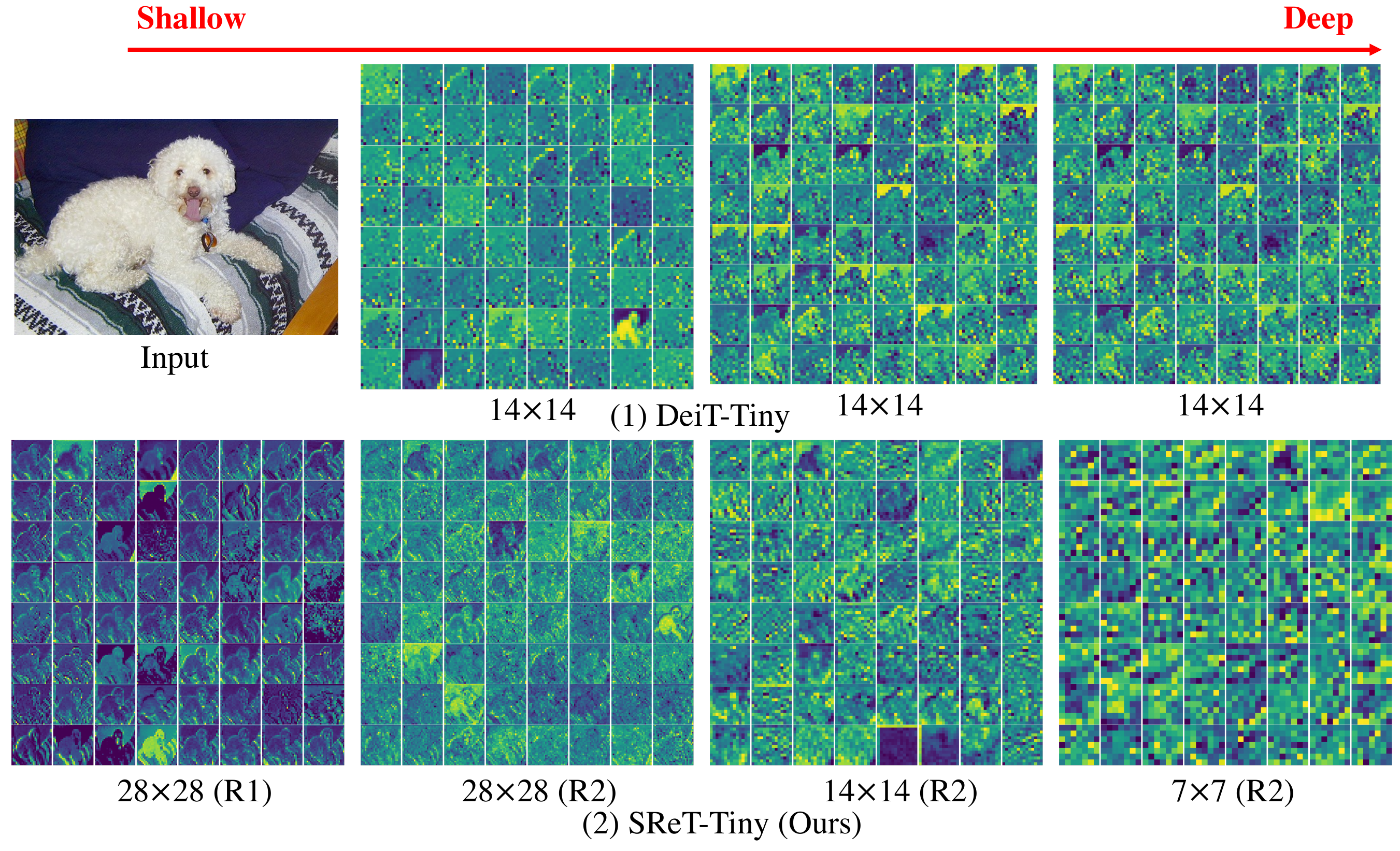}
		\end{minipage}\hfill
		\begin{minipage}[c]{0.4\textwidth}\vspace{0.15in}
			\caption{Illustration of activation distributions on shallow, middle and deep layers of \texttt{DeiT-Tiny} and our \texttt{SReT-T} networks. Under each subfigure, $14\times14$, $28\times28$ and $7\times7$ are the resolutions of feature maps. ``R1/2'' indicates the index of recursive operations in each block.}
			\label{fig:feature_maps_vis}
		\end{minipage}  \vspace{-0.03in}
	\end{figure}

	\noindent{\textbf{Learned response maps.}}
	We visualize the activation maps of DeiT-T and our \texttt{SReT-T} model at shallow and deep layers. As shown in Fig.~\ref{fig:feature_maps_vis}, DeiT is a network with uniform resolution of feature maps (14$\times$14). While, our spatial pyramid structure has different sizes of feature maps along with the depth of the network, i.e., the resolution of feature maps decreases when the depth increases. More interesting observations are discussed in Appendix.
	
	\section{Conclusion}
	
	It is worthwhile considering how to improve the efficiency of parameter utilization for a vision transformer with minimum overhead. In this work, we have summarized and explained several behaviors observed while training such networks. We focused on building an efficient vision transformer with a compact model size through the recursive operation, and the proposed group self-attention approximation method allows us to train in a more efficient manner with recursive transformers. We highlight such a training scheme has not been well-explored yet in previous literature. 
	We attributed the superior performance of sliced recursive transformer to its ability of intensifying the representation quality of intermediate features. 
	We conducted comprehensive experiments to establish the success of our method on the image classification and neural machine translation tasks, not just verifying it in the vision domain, but proving the capability to generalize for multiple modalities and architectures, such as MLP-Mixer.
	
		\appendix
	
	\section*{\Large{Appendix}}
	
	In this appendix, we provide details omitted in the main text, including:
	
	• Section~\ref{flops}: Proof for equivalency of global self-attention and sliced group self-attention with recursive operation on FLOPs. (Sec.~4 ``Approximating Global Self-Attention via Permutation of Group/Local Self-Attentions'' of the main paper.)
	
	• Section~\ref{INV2}: Results of SReT on ImageNet ReaL~\cite{beyer2020we} and ImageNetV2~\cite{recht2019imagenet} datasets. (Sec.~5 ``Experiments and Analysis'' of the main paper.)
	
	• Section~\ref{different_groups}: More ablation results on different permutation designs and numbers of groups when approximating global self-attention on ImageNet-1K. (Sec.~5.3 ``Ablation Studies'' of the main paper.)
	
	• Section~\ref{pseudocode}: Pseudocode for implementing sliced group self-attention. (Sec.~4 ``Approximating Global Self-Attention via Permutation of Group/Local Self-Attentions'' of the main paper.)
	
	• Section~\ref{training_details}: Implementation details of training on ImageNet-1K. (Sec.~5.1 ``Datasets and Experimental Settings'' of the main paper.)
	
	• Section~\ref{details_language}: Hyper-parameters setting for training language models on WMT14 En-De and IWSLT14 De-En datasets. (Sec.~5.1 ``Datasets and Experimental Settings'' and Sec.~5.6 ``Neural Machine Translation'' of the main paper.)
	
	• Section~\ref{details}: Details of our \texttt{SReT-T}, \texttt{SReT-TL}, \texttt{SReT-S} and \texttt{SReT-B} architectures. (Sec.~3 ``Recursive Transformer'' and Sec.~5.3. ``Ablation Studies'' of the main paper.)
	
	• Section~\ref{all_MLP}: Details of All-MLP structure. (Sec.~5.5 ``All-MLP Architecture'' of the main paper.)
	
	• Section~\ref{ablation_LRC}: Ablation study on different LRC designs. (Sec.~3 ``Recursive Transformer'' and Sec.~5.8 ``Analysis and Understanding'' of the main paper.)
	
	• Section~\ref{obs_res}: Observations of Response Maps. (Sec.~5.8 ``Analysis and Understanding'' of the main paper.)
	
	• Section~\ref{more_LRC_SReT}: More evolution visualization of LRC coefficients on ImageNet-1K dataset. (Sec.~5.8 ``Analysis and Understanding'' of the main paper.)
	
	• Section~\ref{vis_language}: Evolution visualization of LRC coefficients in language model on WMT14 En-De dataset. (Sec.~5.6 ``Neural Machine Translation'' and Sec.~5.8 ``Analysis and Understanding'' of the main paper.)
	
	• Section~\ref{more_ablation_extend}: More ablation results on directly expanding the depth of baseline DeiT model on ImageNet-1K dataset. (Sec.~5.8 ``Analysis and Understanding'' of the main paper.)
	
	• Section~\ref{more_exp}: More definitions of ``Feed-forward Networks, Recurrent Neural Networks and Recursive Neural Networks'' and explanations of difference to prior arts. (Sec.~1 ``Introduction'' and Sec.~2 ``Related Work'' of the main paper.)
	
	\section{FLOPs Analysis}\label{flops}
	
	One of the key benefits of our \texttt{SReT} is to control the complexity of a recursive network. We analyze the FLOPs of global (i.e., original) and sliced group self-attentions and compare them with different circumstances of groups in a vision transformer. In this section, we provide a proof to Theorem~1 which we restate below.
	
	{\bf Theorem 1.}
	\  ({\em {\bf \em Equivalency of global self-attention and group self-attention with recursive operation on FLOPs.}) { Let $\{\textbf{N}_{\ell}, \textbf{G}_{\ell}\}\in \mathbb{R}^1$, when  $\textbf{N}_{\ell}=\textbf{G}_{\ell}$,  $\textbf{FLOPs}$(1 $\textbf{V-SA}$) =  $\textbf{FLOPs}$($\textbf{N}_{\ell}\times$ Recursive with $\textbf{G}_{\ell}\times$ $\textbf{G-SAs}$)}.
		The complexity of regular and group self-attentions can be calculated as: (For simplicity, here we assume \#groups and vector dimensions in each recursive operation are the same.)
		\begin{equation} \label{trans000}
			{\bm {\textbf{C}_\text{G-SA}} = \frac{\textbf{N}_{\ell}}{\textbf{G}_{\ell}}\times \bm {\textbf{C}_\text{V-SA} }}
		\end{equation}
		where $\textbf{N}_{\ell}$ is the number of recursive operation and $\textbf{G}_{\ell}$ is the number of group self-attentions in layer $\ell$, i.e., $\ell$-th recursive block. $\textbf{V-SA}$ and $\textbf{G-SA}$ represent the vanilla and group self-attentions, respectively.}
	\begin{proof}
		{\em (Theorem}~1)
		The complexity {\em $\textbf{C}$} of regular self-attention can be calculated as: 
		{\em
			\begin{equation} \label{trans111}
				{\bm {\textbf{C}_\text{V-SA}} = \mathcal{O}(\textbf{L}_\ell^2 \times \textbf{D}_\ell)} 
			\end{equation}
		}
		where {\em $\textbf{L}_\ell$} is the sequence length and {\em $\textbf{D}_\ell$} is the dimensionality of the latent representations.
		
		The complexity  of simple recursive operation without group will be:
		{\em 
			\begin{equation} \label{trans222}
				{\bm {\textbf{C}_\text{recursive}} = \mathcal{O}(\textbf{N}_\ell \times \textbf{L}_\ell^2 \times \textbf{D}_\ell)} 
			\end{equation}
		}
		where {\em $\textbf{N}_\ell$} is the number of recursive operation.
		
		The complexity of sliced group self-attentions with a recursive block can be calculated as: 
		{\em 
			\begin{equation} \label{dif}
				\begin{aligned}
					\bm {\textbf{C}_\text{G-SA} }= \mathcal{O}(\sum_i^{\textbf{N}_\ell}(\textbf{g}_\ell^i \times (\frac{\textbf{L}_\ell}{\textbf{g}_\ell^i})^2 \times \textbf{d}_\ell^i)) \\ =  \mathcal{O}(\sum_i^{\textbf{N}_\ell}(\frac{\textbf{L}_\ell^2}{\textbf{g}_\ell^i} \times \textbf{d}_\ell^i)) 
				\end{aligned}
			\end{equation}
		}
		where {\em $\textbf{g}_\ell^i \in \{\textbf{G}_\ell\}$}, {\em $ \textbf{d}_\ell^i \in \{\textbf{D}_\ell\}$}, $i=1,\dots,\textbf{\em N}_\ell$.
		
		Consider the condition of \#groups $\textbf{\em g}_\ell^i$ and vector dimension $\textbf{\em d}_\ell^i$ in each recursive operation are the same. The complexity of group self-attentions can be re-formulated as: 
		{\em
			\begin{equation} \label{trans333}
				{\bm {\textbf{C}_\text{G-SA} }= \mathcal{O}(\textbf{N}_\ell \times \frac{\textbf{L}_\ell^2}{\textbf{G}_\ell} \times \textbf{D}_\ell) = \frac{\textbf{N}_\ell}{\textbf{G}_\ell}\times \bm {\textbf{C}_\text{V-SA} }}
			\end{equation}
		}
		where $\textbf{\em G}_\ell$ is the number of group self-attentions. When {\em ${ \textbf{N}_\ell= \textbf{G}_\ell}$}, {\em ${\textbf{C}_\text{V-SA} = \textbf{C}_\text{G-SA}}$} and if {\em $ {\textbf{N}_\ell\!<\! \textbf{G}_\ell}$}, {\em ${ \textbf{C}_\text{G-SA}\!<\!\textbf{C}_\text{V-SA}}$}.
	\end{proof}
	
	\section{More Results and Comparisons on ImageNet ReaL~\cite{beyer2020we} and ImageNetV2~\cite{recht2019imagenet} Datasets} \label{INV2}
	
	In this section, we provide results on ImageNet ReaL~\cite{beyer2020we} and ImageNetV2~\cite{recht2019imagenet} datasets. On ImageNetV2~\cite{recht2019imagenet}, we verify our SReT models on three metrics ``Top-Images'', ``Matched Frequency'', and ``Threshold 0.7''. The results are shown in Table~\ref{tab:my-table_INV2}, we achieve consistent improvement over  DeiT  on various network architectures.
	
	\begin{table*}[h]
		\centering \vspace{-0.25in}
		\caption{More Comparison of \texttt{SReT} on ReaL~\cite{beyer2020we} and ImageNetV2~\cite{recht2019imagenet} datasets.}
		\label{tab:my-table_INV2}
		\resizebox{.98\textwidth}{!}{%
			\begin{tabular}{lcccccccc}
				\toprule[1.1pt]
				\multirow{2}{*}{Method} & \multirow{2}{*}{Network} & \multirow{2}{*}{\#Parames} & \multirow{2}{*}{FLOPs} & \multirow{2}{*}{ImageNet} & \multirow{2}{*}{ReaL} & ImageNetV2  & ImageNetV2 & ImageNetV2   \\ 
				& & &     &   &       & Top-images & Matched-frequency & Threshold-0.7 \\ \hline
				DeiT~\cite{touvron2020training} & Tiny & 5.7 & 1.3 &   72.2  &  80.1 & 74.4  &  59.9  & 68.5 \\ 
				\bf SReT &  Tiny  & \bf 4.8 & \bf 1.1 & \bf 76.0 & \bf 83.1  &  \bf 77.9  &  \bf 64.0  & \bf 72.8  \\ \hline
				DeiT~\cite{touvron2020training} & Tiny+Distill & 5.7 & 1.3 &   74.5  &  82.1 & 77.0  &  62.3 & 71.1   \\ 
				\bf SReT &  Tiny+Distill  &   \bf 4.8  & \bf 1.1 &  \bf 77.6  &  \bf  84.4 & \bf 79.6  & \bf 65.7  &  \bf 74.2\\ \hline
				DeiT~\cite{touvron2020training} & Small & 22.1  &  4.6 & 79.8 &  85.7  &  81.0 & 68.1 &  76.4 \\ 
				\bf SReT  & Small     &   \bf 20.9  & \bf 4.2   &  \bf 81.9  &  \bf 86.7 & \bf 82.8 & \bf 70.3 & \bf 78.1  \\ \hline
				DeiT~\cite{touvron2020training} & Small+Distill & 22.1  &  4.6 & 81.2 &  86.8  &  82.5 & 69.7 & 77.5 \\ 
				\bf SReT  & Small+Distill    &   \bf 20.9  & \bf 4.2   &  \bf 82.7 &  \bf 88.1  & \bf 84.0 & \bf 72.3 & \bf 79.9  \\ 
				\bottomrule[1.1pt] 
		\end{tabular}} \vspace{-0.25in}
	\end{table*}
	
	\section{Ablation Results on Different Permutation Designs and Groups Numbers} \label{different_groups}
	
\begin{figure}[t]
		\centering\small
		\includegraphics[width=0.48\linewidth]{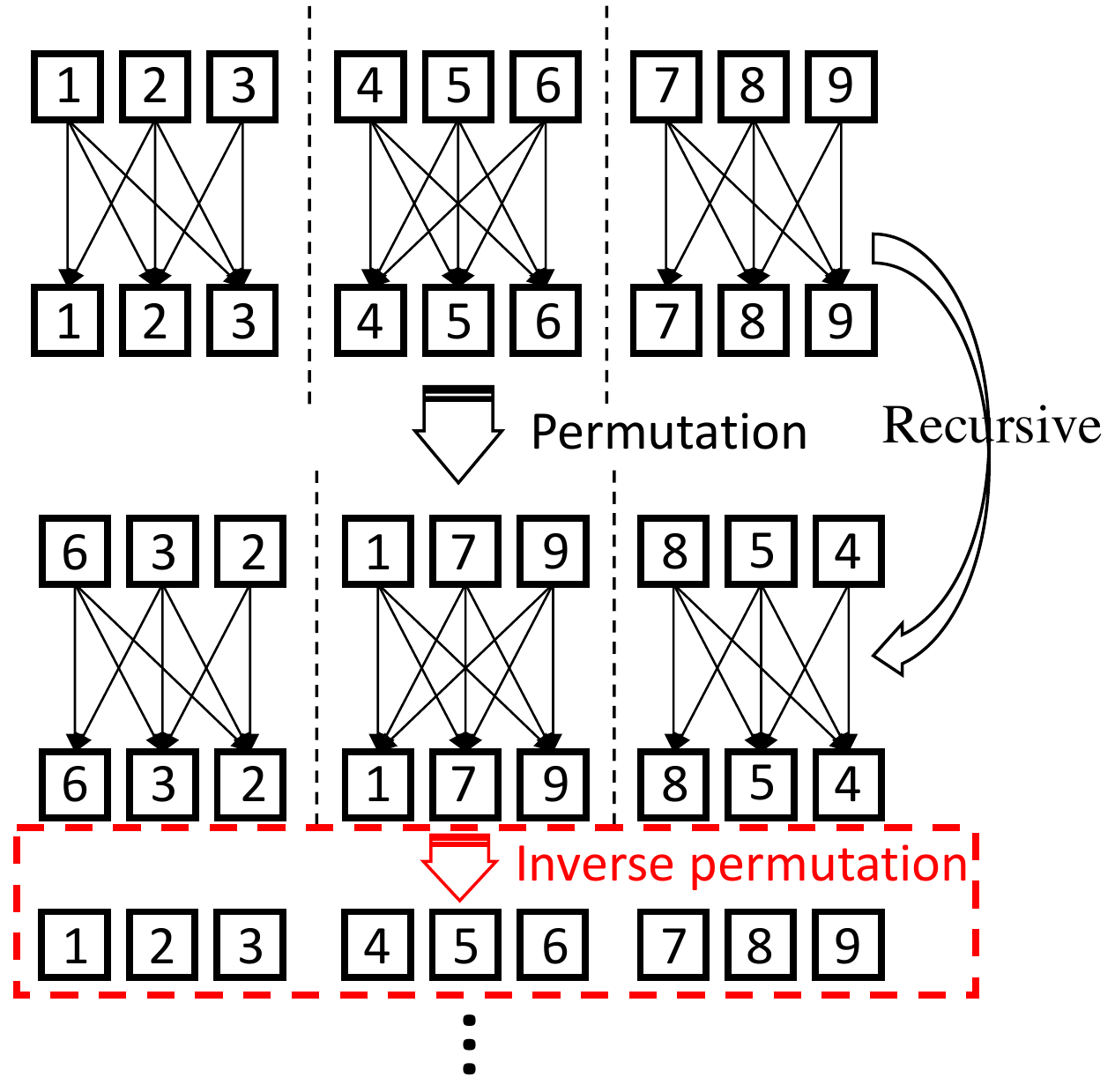} \vspace{-0.05in}
		\caption{Details of group self-attention with permutation designs.}
		\label{fig:expr:inverse} \vspace{-0.15in}
	\end{figure}

	In this section, we explore the different permutation designs and the principle of choosing group numbers for the best accuracy-FLOPs trade-off. We propose to insert an inverse permutation layer to preserve the input order information after the sliced group self-attention operation. The formulation of this operation is shown in Fig.~\ref{fig:expr:inverse} and the ablation results for this design are given in Table~\ref{tab:more_abl_different_groups_inverse} of the first group. In the table, ``P'' represents the permutation layer, ``I'' represents the inverse permutation layer and ``L'' indicates that we did not involve permutation and inverse permutation in {\em the last stage} of models when the number of groups equals 1. We use {\texttt{SReT-T}} and {\texttt{SReT-TL}} as the base structures for the ablation of different groups. In the {\bf Groups} column of the table, we applied two loops of recursion in each recursive block according to the ablation study in Table 1 of our main text. In each pair of the square brackets, the values denote the number of groups for each recursion, and each pair of square brackets represents one stage of blocks in the spatial pyramid based backbone network. We use [8,2][4,1][1,1] as our final \texttt{SReT} structure design since it has the best trade-off on accuracy and computational cost.
	
	\begin{table*}[h]
		\setlength{\tabcolsep}{6pt}
		\centering
		\caption{Ablation results of \texttt{SReT-T} and \texttt{SReT-TL} with different group designs. }
		\label{tab:more_abl_different_groups_inverse}
		\resizebox{.99\textwidth}{!}{%
			\begin{tabular}{c|c|c|c|c|c}
				\toprule[1.1pt]
				\bf Groups   &  \bf Net  & \bf Layers & \bf Params (M) & \bf \#FLOPs (B) & \bf Top-1 (\%) \\ \hline
				[8,8][4,4][1,1] &P  &  20  &  4.99  & 1.08 &    75.41   \\ \hline
				[8,8][4,4][1,1] &P+I  &  20  &  4.99  & 1.08 &   75.94 \\ \hline
				[8,8][4,4][1,1] &P+I-L  &  20  &  4.99  & 1.08 &   \bf 76.06 \\ 
				\bottomrule[0.8pt] 
				\toprule[0.8pt]
				[1,1][1,1][1,1] &   SReT-$\divideontimes$T  &  20  & 4.76  & 1.38 & 76.07 \\ \hline
				[8,8][4,4][1,1] &   SReT-T &  20  & 4.76  & 1.03 & 75.73 \\ \hline
				[16,2][4,2][1,1] &  SReT-T & 20  & 4.76  & 1.01 &  75.79\\ \hline
				[8,2][4,1][1,1]  &  SReT-T & 20  &  4.76 & 1.12  &   75.97  \\ 
				\bottomrule[0.8pt] 
				\toprule[0.8pt]
				[1,1][1,1][1,1]  & SReT-$\divideontimes$TL & 20 & 4.99 & 1.43 &  76.78   \\ \hline
				[8,8][4,4][1,1]  & SReT-TL & 20 & 4.99 & 1.08 &  76.06   \\ \hline
				[8,4][4,2][1,1]  & SReT-TL & 20 & 4.99 & 1.14 &  76.16       \\ \hline
				[8,2][4,1][1,1]  & SReT-TL & 20 & 4.99 & 1.18 & 76.65   \\ \hline
				[8,1][4,1][1,1]  & SReT-TL & 20 & 4.99 & 1.25 & 76.72   \\ \hline
				[16,1][14,1][1,1]  & SReT-TL & 20 & 4.99 & 1.24 & 76.56 \\ \hline
				[49,1][28,1][1,1]& SReT-TL & 20 & 4.99 & 1.23 & 76.30   \\ 
				\bottomrule[1.1pt] 
			\end{tabular}
		} \vspace{-0.1in}
	\end{table*}

	\section{Pseudocode for Sliced Group Self-attention} \label{pseudocode}
	
	The PyTorch pseudocode for implementation of our sliced group self-attention is shown in Algorithm~\ref{alg:code}.

	\begin{algorithm}[t] 
		\caption{PyTorch-like Code for Sliced Group MHSA with 2$\times$ Recursion.}
		\label{alg:code}
		\definecolor{codeblue}{rgb}{0.25,0.5,0.5}
		\lstset{
			backgroundcolor=\color{white},
			basicstyle=\fontsize{7.2pt}{7.2pt}\ttfamily\selectfont,
			columns=fullflexible,
			breaklines=true,
			captionpos=b,
			commentstyle=\fontsize{7.2pt}{7.2pt}\color{codeblue},
			keywordstyle=\fontsize{7.2pt}{7.2pt},
			moredelim=**[is][\color{red}]{@}{@},
		}
		\begin{lstlisting}[language=python,,]
			# num_groups1 and num_groups2: numbers of groups in different recursions
			# recursion: recursive indicator
			
			class SG_Attention(nn.Module):
			def __init__(self, dim, num_groups1=8, num_groups2=4, num_heads=8, qkv_bias=False, qk_scale=None, attn_drop=0., proj_drop=0.):
			super().__init__()
			self.num_heads = num_heads
			# numbers of groups in different recursions
			self.num_groups1 = num_groups1 
			self.num_groups2 = num_groups2
			head_dim = dim // num_heads
			self.scale = qk_scale or head_dim ** -0.5
			
			self.qkv = nn.Linear(dim, dim * 3, bias=qkv_bias)
			self.attn_drop = nn.Dropout(attn_drop)
			self.proj = nn.Linear(dim, dim)
			self.proj_drop = nn.Dropout(proj_drop)
			
			def forward(self, x, recursion):
			B, N, C = x.shape
			if recursion == False:
			num_groups = self.num_groups1
			else:
			num_groups = self.num_groups2
			# we will not do permutation and inverse permutation if #group=1
			if num_groups != 1:
			idx = torch.randperm(N)
			# perform permutation
			x = x[:,idx,:]
			# prepare for inverse permutation
			inverse = torch.argsort(idx)
			
			qkv = self.qkv(x).reshape(B, num_groups, N // num_groups, 3, self.num_heads, C // self.num_heads).permute(3, 0, 1, 4, 2, 5) 
			q, k, v = qkv[0], qkv[1], qkv[2]   # make torchscript happy (cannot use tensor as tuple)
			
			attn = (q @ k.transpose(-2, -1)) * self.scale
			attn = attn.softmax(dim=-1)
			attn = self.attn_drop(attn)
			
			x = (attn @ v).transpose(2, 3).reshape(B, num_groups, N // num_groups, C)
			x = x.permute(0, 3, 1, 2).reshape(B, C, N).transpose(1, 2)
			if recursion == True and num_groups != 1:
			# perform inverse permutation 
			x = x[:,inverse,:]
			x = self.proj(x)
			x = self.proj_drop(x)
			return x
			...
		\end{lstlisting} 
	\end{algorithm} 
	
	\section{Training Details on ImageNet-1K} \label{training_details}
	
	On ImageNet-1K, we conduct experiments on three training schemes: (1) conventional training with one-hot labels; (2) distillation with soft labels from a pre-trained teacher; (3) finetuning from distilled parameters with higher resolution. Our training settings and hyper-parameters mainly follow the designs of DeiT~\cite{touvron2020training}. A detailed introduction of these settings is shown in Table~\ref{tab:my-table_conventional_cokpare},~\ref{tab:my-table_distillation} and~\ref{tab:my-table_finetuning} with an item-by-item comparison. 
	
	\begin{table}[t]
		\centering
		\begin{minipage}{0.43\textwidth}
			\centering
			\caption{Hyper-parameter details of conventional training.}
			\label{tab:my-table_conventional_cokpare}
			\resizebox{.95\textwidth}{!}{%
				\begin{tabular}{lccc}
					\toprule[1.1pt]
					Method          & SReT-T    & SReT-TL    & SReT-S      \\
					Epoch           & 300     & 300  & 300         \\
					Batch size      & 1024    & 1024 & \bf 512       \\
					Optimizer       & AdamW   & AdamW  & AdamW       \\
					Learning rate   & 0.001   & 0.001   & 0.001       \\
					Weight decay    & 0.05    & 0.05  & 0.05        \\
					Warmup epochs   & 5       & 5     &  5        \\
					Label smoothing & 0.1     & 0.1   & 0.1        \\
					Stoch. Depth & 0.1     & 0.1   & \bf 0.2        \\
					\bottomrule[1.1pt] 
				\end{tabular}
			}
		\end{minipage}
		%
		\hspace{0.1in}
		\begin{minipage}{0.47\textwidth}
			\centering
			\small
			\setlength{\tabcolsep}{2pt}
			\centering
			\caption{Hyper-parameter  details of soft distillation training.}
			\label{tab:my-table_distillation}
			\resizebox{.99\textwidth}{!}{%
				\begin{tabular}{lcc}
					\toprule[1.1pt]
					Method          & DeiT    & SReT         \\
					\bf Label      & \bf one-hot+hard distillation    &\bf soft distillation         \\
					Epoch           & 300     & 300         \\
					Batch size      & 1024    & 1024         \\
					Optimizer       & AdamW   & AdamW       \\
					Learning rate   & 0.001   & 0.001        \\
					\bottomrule[1.1pt] 
				\end{tabular}
			}
			\centering
			\caption{Hyper-parameter details of higher-resolution finetuning.}
			\label{tab:my-table_finetuning}
			\resizebox{.6\textwidth}{!}{%
				\begin{tabular}{lcc}
					\toprule[1.1pt]
					Method          & DeiT    & SReT         \\
					Resolution      & 384   & 384         \\
					\bf Weight decay    & \bf 1e-8    &\bf 0.0        \\
					Learning rate   & 5e-6   & 5e-6        \\
					\bottomrule[1.1pt] 
				\end{tabular}
			}
		\end{minipage} %
	\end{table}
	
	\noindent{\textbf{Conventional Training from Scratch with One-hot Label.}} As shown in Table~\ref{tab:my-table_conventional_cokpare}, we use batch-sizes of 512/1024 for training our models and the default initial learning rate is $1e$-3, while from our experiments, larger initial {\em lr} of $2e$-3 with more warmup epochs of 30 can favorably improve the accuracy. Other settings are following~\cite{touvron2020training}.
	
	\begin{figure*}[h]
		\centering \vspace{-0.01in}
		\includegraphics[width=0.99\linewidth]{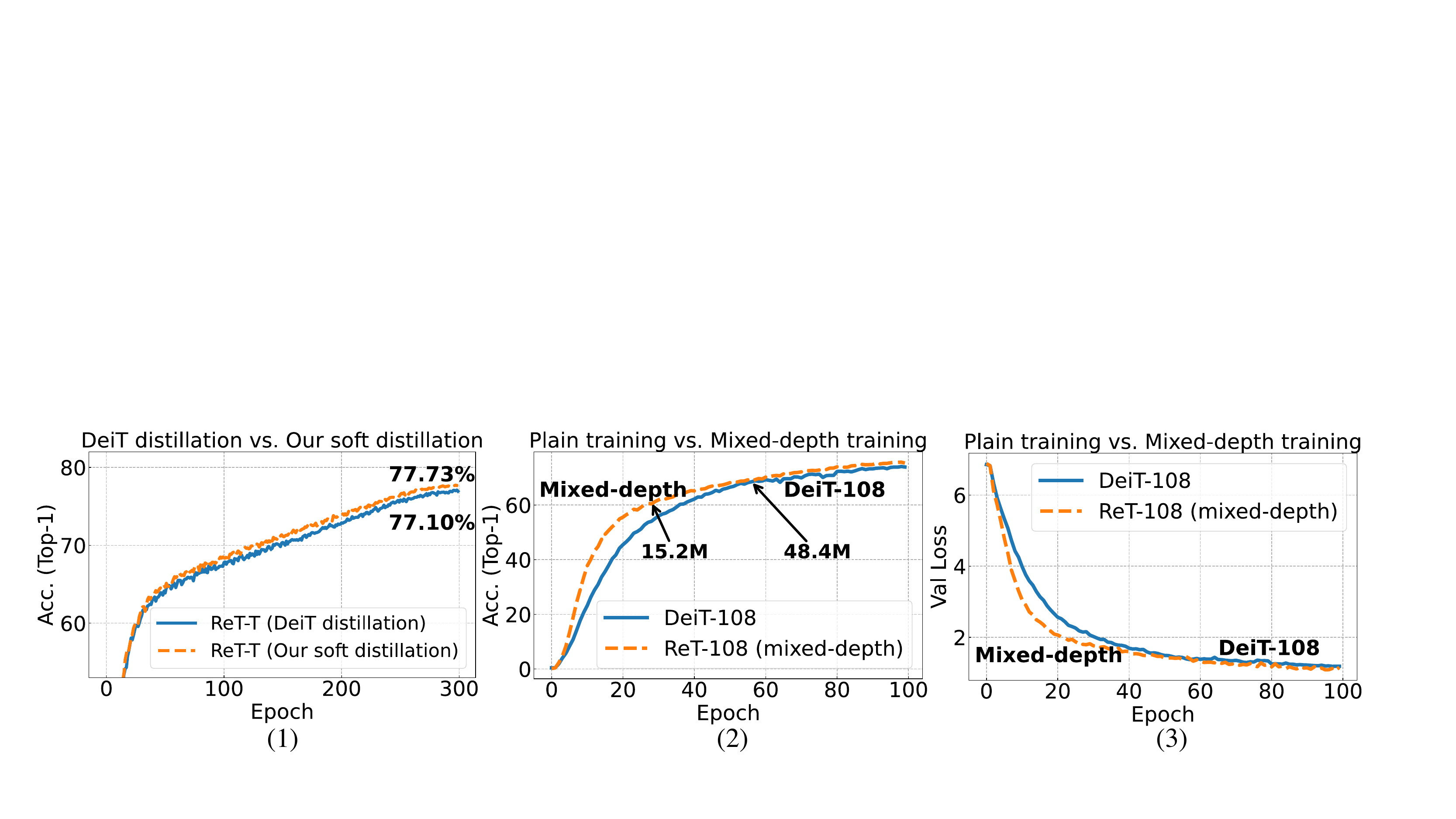} \vspace{-0.1in}
		\caption{Comprehensive ablation study on different design factors.}
		\label{fig:ablation_all} \vspace{-0.1in}
	\end{figure*}
	
	\noindent{\textbf{Distillation Strategy.}} Knowledge distillation~\cite{hinton2015distilling} is a popular way to boost the performance of a student network. Recently, many promising results~\cite{pham2020meta,shen2020meal,xie2020self} have been achieved using this technique. On vision transformer, DeiT~\cite{touvron2020training} proposed to distill tokens together with hard predictions from the teacher, and it claimed that using one-hot label with hard distillation can achieve the best accuracy. This seems counterintuitive since soft labels can provide more subtle differences and fine-grained information of the input. In this work, through a proper distillation scheme, our soft label based distillation framework (one-hot label is not used) consistently obtained better performance than DeiT. 
	Our loss is a soft version of cross-entropy between teacher and student's outputs as used in~\cite{romero2014fitnets,bagherinezhad2018label,shen2021is}:
	\begin{equation}
		{{\mathcal{L}}_{CE}}({\mathcal{S}_\mathbf{W} }) =  - \frac{1}{N}{\sum\limits_{i = 1}^N {{\bf P}_{{\mathcal{T}_\mathbf{W} }}({\mathbf{z}})\log } } {\bf P}_{{\mathcal{S}_\mathbf{W} }}({\mathbf{z}})
	\end{equation}
	where ${\bf P}_{\mathcal{T}_\mathbf{W}}$ and ${\bf P}_{\mathcal{S}_\mathbf{W}}$ are the outputs of teacher and student, respectively.

	\noindent{\textbf{Distillation from Scratch.}} As shown in Table~\ref{tab:my-table_distillation}, we use soft predictions solely from RegNetY-16GF~\cite{radosavovic2020designing} as a teacher instead of {\em one-hot label + hard distillation} used in~\cite{touvron2020training}. The ablation study on this point is provided in Fig.~\ref{fig:ablation_all} (1) with \texttt{SReT-T}.
	
	\noindent{\textbf{Spatial Pyramid (SP) Design.}} Pyramids~\cite{1641019,he2015spatial} are an effective design in conventional vision tasks. The resolution of the shallow stage in a network is usually large, SP can help to redistribute the computation from shallow to deep stages of a network according to their representation ability. Here, we follow the construction principles~\cite{heo2021rethinking} but replacing the first patch embedding layer with a {\em Stem block} (i.e., a stack of three 3$\times$3 convolution layers with stride = 2) following~\cite{shen2017dsod}. 
	
	\begin{wrapfigure}{r}{5.3cm}
	\vspace{-0.32in}
	\includegraphics[width=0.99\linewidth]{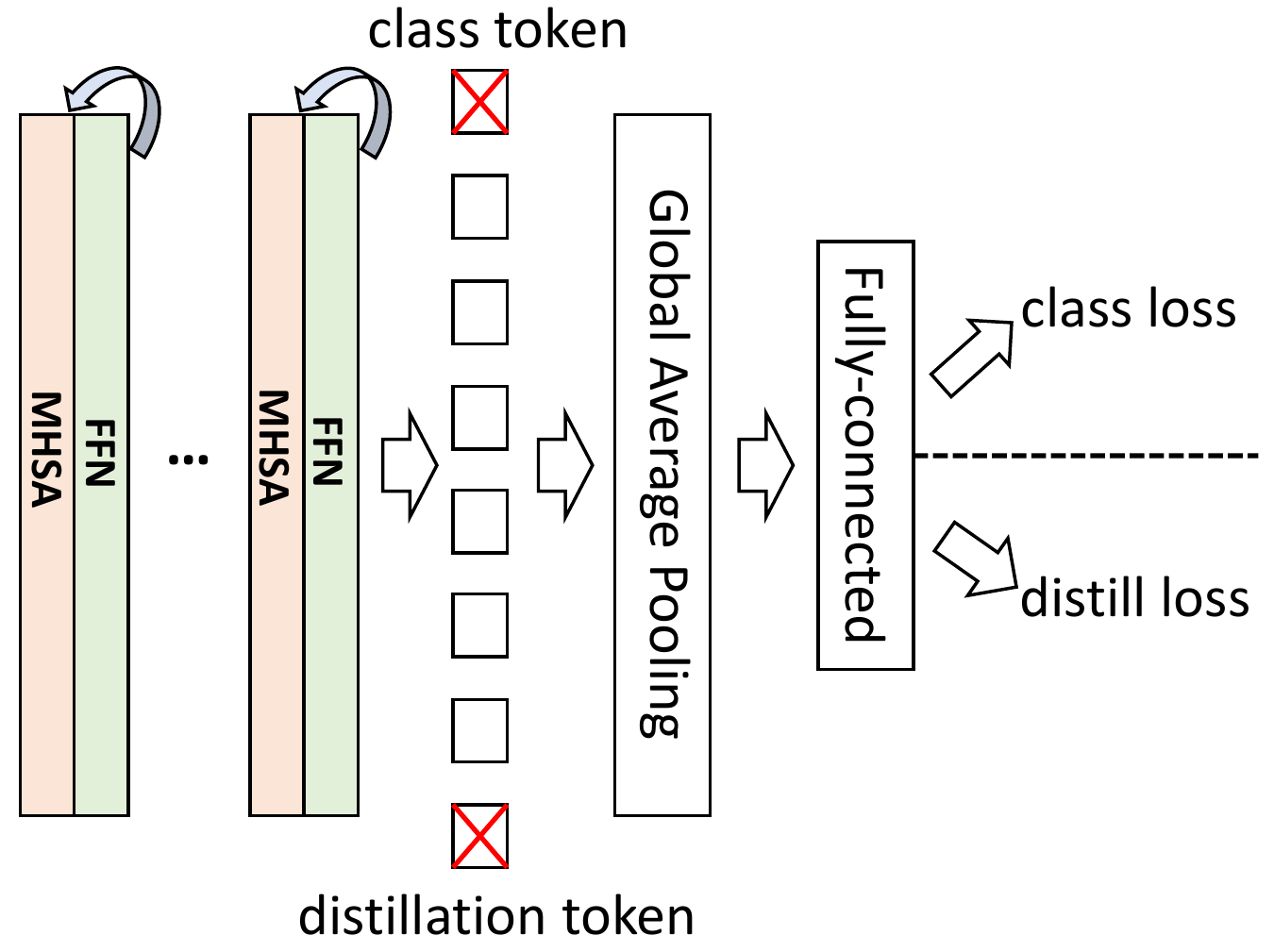}
	\vspace{-0.1in}
	\caption{Our modifications by removing class token and distillation token.}
	\label{fig:minor_modification}
	\vspace{-0.3in}
	\end{wrapfigure}
	
	\noindent{\textbf{Other Small Modifications.}} Considering the unique properties of vision modality compared to the language, we further apply some minor modifications on our network design, some of them have been proven useful on CNNs in the vision domain, including: (i) We remove the class token and replace with a {\em global average pooling (GAP)} on the last output together with a fully-connected layer; (ii) We also remove the distillation token if the training process involves KD, which means we use the same feature embedding for both the ground-truth labels in standard training, and distillation with soft labels from the teacher. (iii) When fine-tuning from low resolution (224$\times$224) to high resolution (384$\times$384)~\cite{touvron2020training}, following the perspective of~\cite{shen2020meal} that to increase the capacity of a model, we do not apply {\em weight decay} (set it as 0) during fine-tuning. Generally, the above modifications can slightly save parameters, boost the performance and significantly improve the simplicity of the whole framework. The illustration of these modifications is shown in Fig.~\ref{fig:minor_modification}.
	
	\begin{figure*}[h]
		\centering \vspace{-0.2in}
		\includegraphics[width=0.98\linewidth]{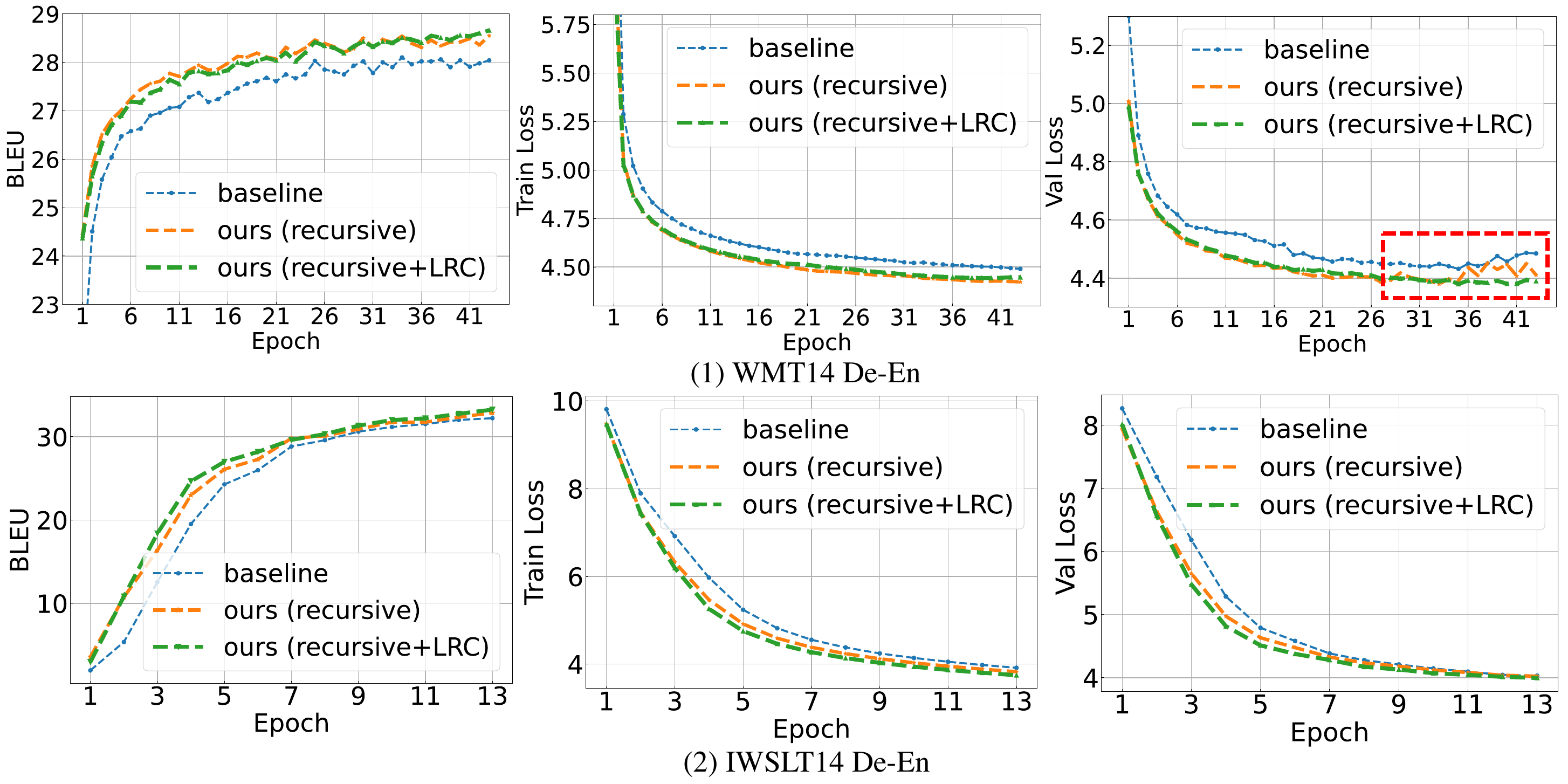}
		\caption{Comparison of BLEU, training loss and val loss on WMT14 En-De (top) and IWSLT14 De-En  datasets (bottom). The red dashed box indicates that LRC makes training more stable.}
		\label{fig:NMT_results_full}
	\end{figure*}
	
	\section{Hyper-parameter Settings of Language Models} \label{details_language}
	We test our proposed method on two public language datasets: IWSLT14 De-En and WMT14 En-De translation tasks. We  describe experimental settings in detail in Table~\ref{tab:my-table_language}.
	
	\noindent{\textbf{Network Configurations.}} We use the Transformer~\cite{vaswani2017attention} implemented in Fairseq~\cite{fair_sque} that shares the decoder input and output embedding as the basic NMT model.
	
	\begin{table*}[t]
		\caption{Training details of our language models. The architectures we used are in Fairseq~\cite{fair_sque}.}
		\centering
		\label{tab:my-table_language}
		\resizebox{.985\textwidth}{!}{%
			\begin{tabular}{lcc}
				\toprule[1.1pt]
				Method                           & IWSLT14 De-En                & WMT14 En-De                  \\ \hline
				arch                             & transformer\_iwslt\_de\_en   & transformer\_wmt\_en\_de     \\
				share decoder input output embed & True                         & True                         \\ \hline
				optimizer                        & Adam                         & Adam                         \\
				adam-betas                       & (0.9, 0.98)                  & (0.9, 0.98)                  \\
				clip-norm                        & 0.0                          & 0.0                          \\ \hline
				learning rate                    & 5e-4                         & 5e-4                         \\
				lr scheduler                     & inverse sqrt                 & inverse sqrt                 \\
				warmup updates                   & 4K                           & 4K                           \\
				dropout                          & 0.3                          & 0.3                          \\
				weight decay                     & 0.0001                       & 0.0001                       \\ \hline
				criterion                        & label smoothed cross-entropy & label smoothed cross-entropy \\
				label smoothing                  & 0.1                          & 0.1                          \\
				max tokens                       & 4096                         & 4096                         \\
				\bottomrule[1.1pt] 
			\end{tabular}
		}
	\end{table*}

	\begin{figure*}[t]
		\centering
		\includegraphics[width=0.82\linewidth]{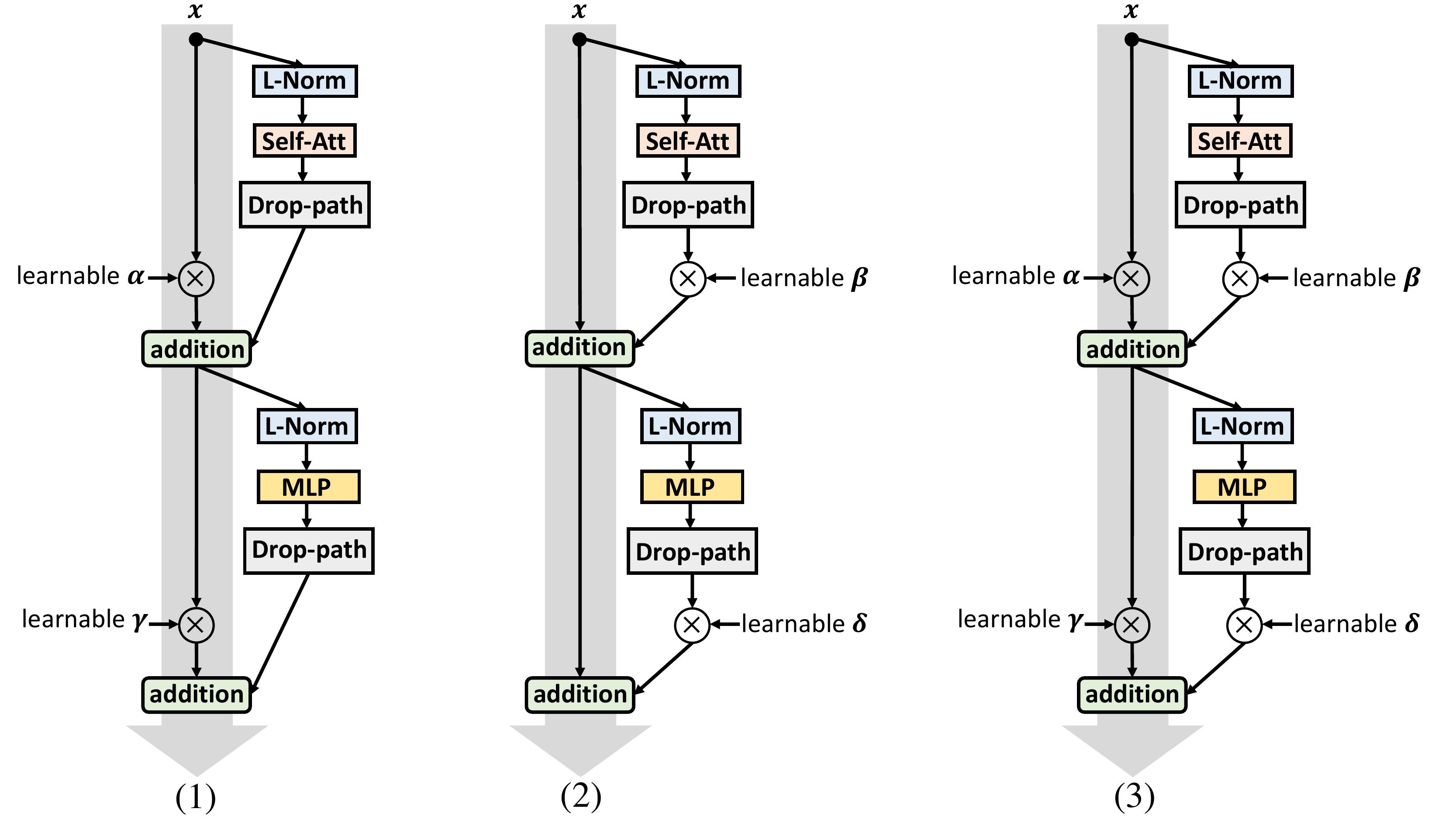}
		\caption{Ablation study on different LRC designs.}
		\label{fig:designs_coefficients}
	\end{figure*}
	
	\section{Details of Our \texttt{SReT} Architectures} \label{details}
	The details of our \texttt{SReT-T}, \texttt{SReT-TL}, \texttt{SReT-S} and \texttt{SReT-B} architectures are shown in Table~\ref{details_network}. In each recursive transformer block $[[.]\times A]\times B$, $A$ is the number of blocks with self-contained (non-shared) parameters, $B$ is the number of recursive operations for each block. For $C\times \text{FFN}$ and $D\times \text{NLL}$, $C$ and $D$ are the dimensions (ratios) of hidden features between the two fully-connected layers.
	
	\renewcommand{\arraystretch}{1.2}
	\begin{table*}[h]
		\centering
		\caption{\texttt{SReT} architectures (Input size is 3$\times224\times224$, sliced group self-attention is not included for simplicity.)} 
		\label{details_network}
		\resizebox{.99\textwidth}{!}{%
			\begin{tabular}{c|c|ccc}
				\toprule[1.1pt]
				\multicolumn{2}{c|}{Layers}                                & Output Size  & SReT-T  & SReT-TL   \\ \hline
				\multirow{3}{*}{Stem}     & Conv-BN-ReLU     &32$\times$112$\times$112  & 3$\times$3 conv, stride 2   & 3$\times$3 conv, stride 2  \\ \cline{2-5}
				& Conv-BN-ReLU & 64$\times$56$\times$56  & 3$\times$3 conv, stride 2  &3$\times$3 conv, stride 2   \\ \cline{2-5}
				& Conv-BN-ReLU & 64$\times$28$\times$28 & 3$\times$3 conv, stride 2  &3$\times$3 conv, stride 2 \\ \cline{1-5}
				\multicolumn{2}{c|}{\begin{tabular}[c]{@{}c@{}}Recursive T Block\\ (1)\end{tabular}} & 64$\times$28$\times$28   & \ret{2}{2}{64}{3.6}{1.0} & \ret{2}{2}{64}{4.0}{1.0}   \\ \hline
				\multicolumn{2}{c|}{\multirow{1}{*}{\begin{tabular}[c]{@{}c@{}}Conv-Pooling Layer (1)\end{tabular}}} & 128$\times$14$\times$14   & 3$\times$3 conv, stride 2, group 64& 3$\times$3 conv, stride 2, group 64\\ \hline
				\multicolumn{2}{c|}{\begin{tabular}[c]{@{}c@{}}Recursive T Block\\ (2)\end{tabular}} & 128$\times$14$\times$14   & \ret{5}{2}{128}{3.6}{1.0} & \ret{5}{2}{128}{4.0}{1.0} \\ \hline
				\multicolumn{2}{c|}{\multirow{1}{*}{\begin{tabular}[c]{@{}c@{}}Conv-Pooling Layer (2)\end{tabular}}} & 256$\times$7$\times$7   & 3$\times$3 conv, stride 2, group 128& 3$\times$3 conv, stride 2, group 128\\ \hline
				\multicolumn{2}{c|}{\begin{tabular}[c]{@{}c@{}}Recursive T Block\\ (3)\end{tabular}} & 256$\times$7$\times$7   & \ret{3}{2}{256}{3.6}{1.0} & \ret{3}{2}{256}{4.0}{1.0} \\ \hline
				\multicolumn{2}{c|}{Global Average Pooling}                           &   256$\times$1$\times$1     & AdaptiveAvgPool &AdaptiveAvgPool  \\ \hline
				\multicolumn{2}{c|}{Linear Layer}                           &    \multicolumn{3}{c}{1000}                                                         \\ \hline
				\multicolumn{2}{c|}{\#Params (M)}                           &   & \multicolumn{1}{c|}{4.8 M}  &  5.0 M                   \\ \hline
				\multicolumn{2}{c|}{Accuracy (\%)}                           &  & \multicolumn{1}{c|}{76.1}  &  76.8              \\ \hline
				\multicolumn{2}{c|}{Distilled Accuracy (\%)}                 &  & \multicolumn{1}{c|}{77.7}  &  77.9          \\ \hline
				\multicolumn{2}{c|}{Finetuning Accuracy $\uparrow$384 (\%)}        &         &   \multicolumn{1}{c|}{79.7}  &  80.0          \\ 
				\bottomrule[1.1pt] 
			\end{tabular}
		}
		\resizebox{.97\textwidth}{!}{%
			\begin{tabular}{c|c|cccc}
				\toprule[1.1pt]
				\multicolumn{2}{c|}{Layers}                                & Output Size  &  SReT-S   & Output Size  &  SReT-B     \\ \hline
				\multirow{3}{*}{Stem}     & Conv-BN-ReLU      & 63$\times$112$\times$112  &  3$\times$3 conv, stride 2   & 96$\times$112$\times$112  &  3$\times$3 conv, stride 2  \\ \cline{2-6}
				& Conv-BN-ReLU  &126$\times$56$\times$56 &3$\times$3 conv, stride 2  &168$\times$56$\times$56 &3$\times$3 conv, stride 2   \\ \cline{2-6}
				& Conv-BN-ReLU  &126$\times$28$\times$28 &3$\times$3 conv, stride 2 & 336$\times$28$\times$28 &3$\times$3 conv, stride 2 \\ \cline{1-6}
				\multicolumn{2}{c|}{\begin{tabular}[c]{@{}c@{}}Recursive T Block\\ (1)\end{tabular}} & 126$\times$28$\times$28   & \ret{2}{2}{126}{3.0}{2.0} & 336$\times$28$\times$28   & \ret{2}{2}{336}{3.0}{2.0} \\ \hline
				\multicolumn{2}{c|}{\multirow{1}{*}{\begin{tabular}[c]{@{}c@{}}Conv-Pooling Layer (1)\end{tabular}}} & 252$\times$14$\times$14   & 3$\times$3 conv, stride 2, group 126 & 672$\times$14$\times$14   & 3$\times$3 conv, stride 2, group 336\\ \hline
				\multicolumn{2}{c|}{\begin{tabular}[c]{@{}c@{}}Recursive T Block\\ (2)\end{tabular}} & 252$\times$14$\times$14   & \ret{5}{2}{252}{3.0}{2.0} & 672$\times$14$\times$14   & \ret{5}{2}{672}{3.0}{2.0} \\ \hline
				\multicolumn{2}{c|}{\multirow{1}{*}{\begin{tabular}[c]{@{}c@{}}Conv-Pooling Layer (2)\end{tabular}}} & 504$\times$7$\times$7   & 3$\times$3 conv, stride 2, group 252 & 1344$\times$7$\times$7   & 3$\times$3 conv, stride 2, group 672\\ \hline
				\multicolumn{2}{c|}{\begin{tabular}[c]{@{}c@{}}Recursive T Block\\ (3)\end{tabular}} & 504$\times$7$\times$7   & \ret{3}{2}{504}{3.0}{2.0} & 1344$\times$7$\times$7   & \ret{3}{2}{1344}{3.0}{2.0}\\ \hline
				\multicolumn{2}{c|}{Global Average Pooling}       &504$\times$1$\times$1  &AdaptiveAvgPool    &1344$\times$1$\times$1  &AdaptiveAvgPool                                      \\ \hline
				\multicolumn{2}{c|}{Linear Layer}                           &    \multicolumn{4}{c}{1000}                                                         \\ \hline
				\multicolumn{2}{c|}{\#Params (M)}                           &   & \multicolumn{1}{c|}{20.9 M}  &    &\multicolumn{1}{c}{71.2 M}                      \\ \hline
				\multicolumn{2}{c|}{Accuracy (\%)}                           &  & \multicolumn{1}{c|}{82.0}  &    &\multicolumn{1}{c}{82.7}               \\ \hline
				\multicolumn{2}{c|}{Distilled Accuracy (\%)}                 &  & \multicolumn{1}{c|}{82.8}  &   &\multicolumn{1}{c}{83.7}           \\ \hline
				\multicolumn{2}{c|}{Finetuning Accuracy $\uparrow$384 (\%)}          &       &   \multicolumn{1}{c|}{83.8}  &  &\multicolumn{1}{c}{84.8}           \\ 
				\bottomrule[1.1pt] 
			\end{tabular}
		} \vspace{-0.1in}
	\end{table*}
	
	\section{All-MLP Structure} \label{all_MLP}
	We use  B/16 in Mixer architectures~\cite{tolstikhin2021mlpmixer} as our backbone network. In particular, it contains 12 layers, the patch resolution is $16\times16$, the hidden size $C$ is 768, the sequence length $S$ is 196, the MLP dimension $D_C$ and $D_S$ are 3072 and 384, respectively.
	
	\section{Ablation Study on Different LRC Designs} \label{ablation_LRC}
	
	In this section, we verify the effectiveness of different LRC designs as shown in Fig.~\ref{fig:designs_coefficients}, including: (1) learnable coefficients on the identity mapping branch; (2) learnable coefficients on the main self-attention/MLP branch; (3) our used design in the main text, i.e., including learnable coefficients on both branches.
	
\begin{figure}[h]
		\centering
		\includegraphics[width=0.88\linewidth]{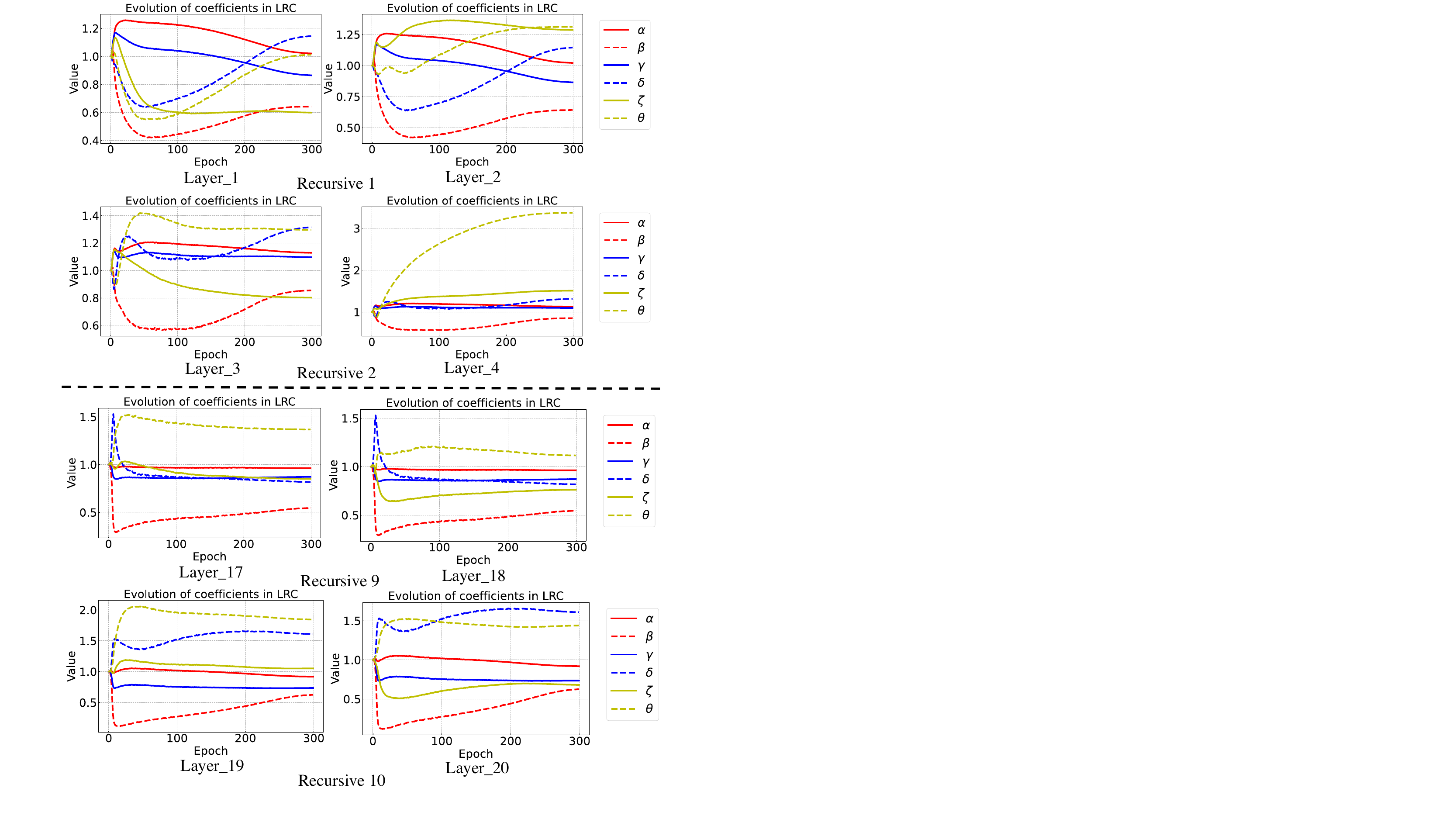}
		\caption{Evolution of coefficients at different recursive blocks and layers.}
		\label{fig:more_vis_coefficients}
	\end{figure}

	The quantitative results of different LRC designs are shown in Table~\ref{tab:LRC}, we can observe that strategy (1) is slightly better than (2), while, (3) can achieve consistent improvement over (1) and (2), and it is applied in our main text. We further visualize more evolution visualizations on various layers/depths of our \texttt{SReT-TL} architecture. The results are shown in Fig.~\ref{fig:more_vis_coefficients} and the analysis is provided in Sec.~\ref{more_LRC_SReT}.
	
	\begin{table}[h]
		\centering
		\caption{Ablation study on different LRC designs.}
		\label{tab:LRC}
		\resizebox{.6\textwidth}{!}{%
			\begin{tabular}{l|c|c}
				\toprule[1.1pt]
				Method         & \#Params (M) & Top-1 Acc. (\%) \\ \hline
				Baseline (\texttt{SReT-TL} w/o LRC)   &    5.0   &  74.7  \\ \hline
				on $x$ branch (1)  &   5.0 & 75.0 \\ 
				on $f$ branch (2) &  5.0 &  74.9\\ \hline
				on both (3) &  5.0  & \bf 75.2   \\ 
				\bottomrule[1.1pt] 
			\end{tabular}
		}
	\end{table}

	\section{Observations of Response Maps} \label{obs_res}
	
	We have a few interesting observations on the visualizations of Fig.~8 (main text): (1) In the uniform size of transformer DeiT, information in the shallow layers is basically vague, blurry and lacks details. In contrast, the high-level layers contain stronger semantic information and are more aligned with the input. However, our model has a completely different behavior: first, in the same block but with different recursive operations, we can observe that the features are hierarchical (in Fig.~8 of main text (2)). Taken as a whole, shallow layers can capture more details like edges, shapes and contours and deep layers focus on the high-level semantic information, which is similar to CNNs. We emphasize such hierarchical representation enabled by recursion and spatial pyramid is critical for vision modality like images.
	
	\vspace{-0.06in}
	\section{More Evolution Visualization of LRC Coefficients on ImageNet-1K Dataset} \label{more_LRC_SReT}
	The visualizations of coefficients evolution at different recursive blocks and layers are shown in Fig.~\ref{fig:designs_coefficients}. Intriguingly, we can observe in the deep layers of recursive blocks, $\alpha$ tends to be one stably during the whole training. Other coefficients on the identity mapping ($\gamma$ and $\zeta$) are holding fixed values that are also close to one during the training. This phenomenon indicates that the identity mapping branch tends to pass the original signal with small scaling. Moreover, it seems the contributions of the two branches have a particular proportion for the particular depth of layers.

	\section{Evolution Visualization of LRC Coefficients on Language Model} \label{vis_language}
	
\begin{figure}[h]
		\centering
		\includegraphics[width=0.50\linewidth]{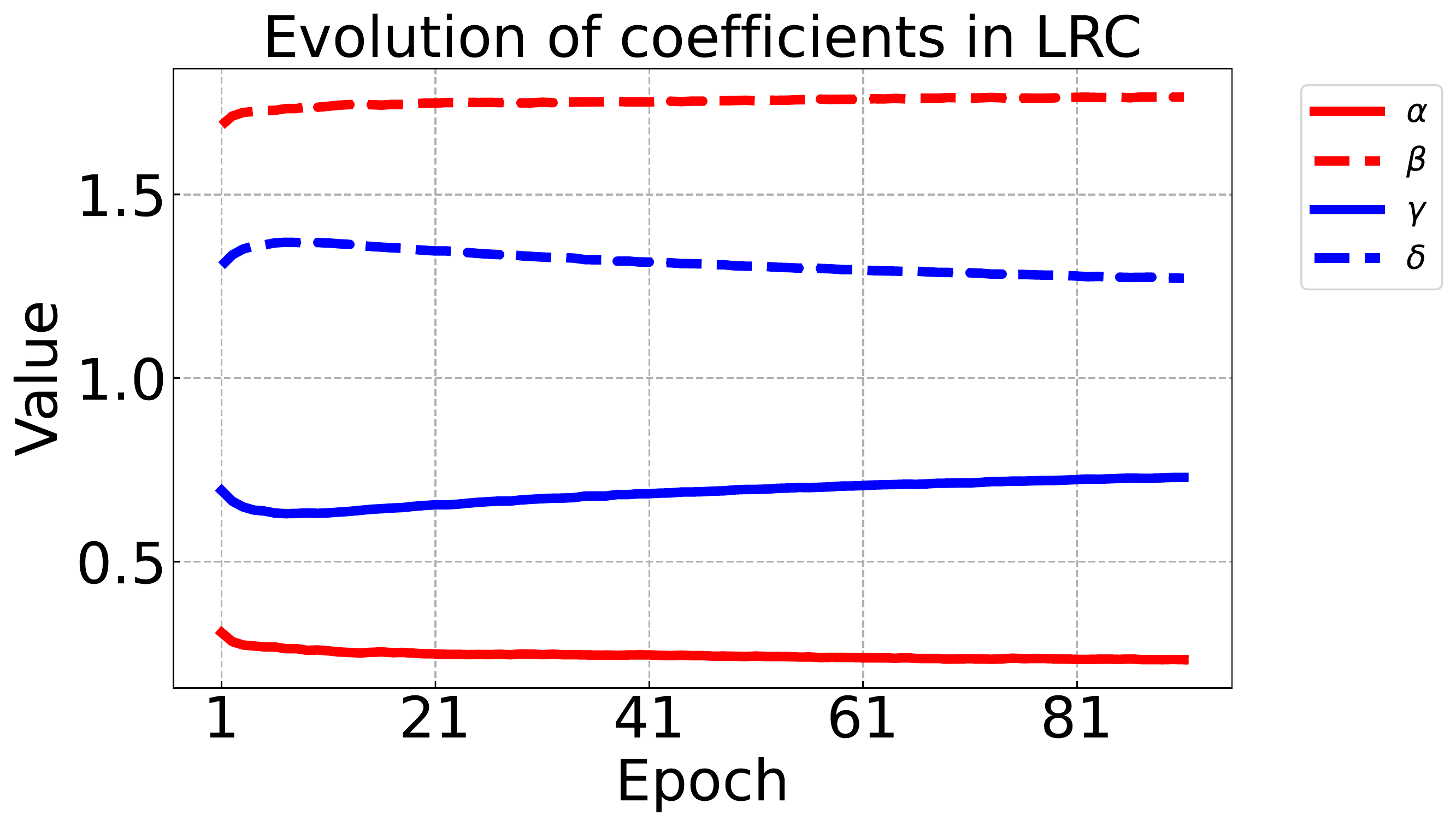}
		\caption{Evolution of coefficients on language of WMT14 En-De dataset.}
		\label{fig:mix_coefficients_language}\vspace{-0.2in}
	\end{figure}

	The visualization of coefficients evolution on the language model is shown in Fig.~\ref{fig:mix_coefficients_language}. Different from the evolution in vision transformer models, the coefficients in language model are more stable during training with small variance. Also, they are symmetrical with value one.
	
	\section{More Ablation Results on Directly Enlarging Depth of Baseline DeiT Model} \label{more_ablation_extend}
	
	In this section, we provide the results by directly expanding the depth of baseline DeiT model, as shown in Table~\ref{tab:more_depth}. We can see deeper na\"ive DeiT could not bring additional gain on performance since the deeper and heavier network is usually more difficult to learn meaningful and diverse intermediate features, while our recursive operation through sharing/reusing parameters is an effective way to enlarge the depth of a transformer, meanwhile, obtaining extra improvement.
	
	\begin{table}[h]
		\centering \vspace{-0.2in}
		\caption{More ablation results on directly expanding depth of baseline DeiT model. * indicates that the total number layers of our network is 20 (recursive transformer blocks) + 10 (NLL) + 3 (image patch embeddings). Permutation and inverse permutation layers are not included.}
		\label{tab:more_depth}
		\resizebox{.62\textwidth}{!}{
			\begin{tabular}{l|c|c|c}
				\toprule[1.1pt]
				Method        & \#Layers & \#Params (M) & Top-1 Acc. (\%) \\ \hline
				DeiT-Tiny~\cite{touvron2020training}  &  12  &    5.7    &  72.20     \\ \hline
				+ extend depth & 24  &    11.55    & 77.35 \\ \hline
				+ extend depth  & 36  &   16.39   &  77.18    \\ \hline
				+ extend depth  & 48 &  21.73  &  75.89        \\ \hline\hline
				Ours (SReT-S) & 33* & 20.90 &   \bf 81.90  \\ 
				\bottomrule[1.1pt] 
			\end{tabular}
		}
	\end{table}
	
	\section{More Definitions and Explanations to Prior Arts} \label{more_exp}
	
	\noindent{\textbf{Feed-forward Networks, Recurrent Neural Networks and Recursive Neural Networks.}} 
	To clarify the definition of proposed recursive operation, we distinct recursive neural networks from feed-forward networks and recurrent neural networks. 
	Feed-forward networks, such as CNNs and transformers, are directed acyclic graphs (DAG). The information path in the feed-forward processing is unidirectional, making the feed-forward networks hard to tackle the structured data with long-span correlations. Recurrent neural networks (RNNs) are usually developed to process the time-series and other sequential data. They output predictions based on the current input and past memory, so they are capable of processing data that contains long-term interdependent compounds like language. Recursive network is a less frequently used term compared to other two counterparts. Recursive refers to repeating or reusing a certain piece of a network. Different from RNNs that repeat the same block throughout the whole network, recursive neural network selectively repeats critical blocks for particular purposes. The recursive transformer iteratively refines its representations for all image patches in the sequence.

	\noindent{\textbf{Difference to Prior Arts: }}On CNNs, ShuffleNet~\cite{zhang2018shufflenet} uses inerratic {\em  shuffle} for efficient design while it is not truly random. Thus, there is no {\em inverse} operation involved. In contrast, our {\em permutation} is entirely stochastic and {\em inverse} is crucial since self-attention is sensitive to tokens' order. The na\"ive group self-attention only has interaction within the window, Swin~\cite{liu2021swin} addresses this using shifted windows across {\em different layers}. While, we solve it by integrating ``slice+permutation+recursion'' on the {\em same layer's parameters}, so each layer enables to interact with all other windows, not across layers as Swin. 

	\clearpage
	%
	%
	\bibliographystyle{splncs04}
	\bibliography{egbib}
\end{document}